\documentclass[journal]{IEEEtran}
\usepackage{graphicx}
\usepackage{booktabs}
\usepackage{makecell}
\usepackage{amsmath}
\usepackage{color}
\usepackage{tabularx}
\usepackage{subfigure}
\usepackage[graphicx]{realboxes}
\usepackage{multirow}

\usepackage[breaklinks=true,bookmarks=true]{hyperref}

\usepackage[table]{xcolor}

\usepackage{orcidlink}

\usepackage{xspace}
\usepackage[mathscr]{euscript}
\usepackage{amssymb}
\makeatletter
\DeclareRobustCommand\onedot{\futurelet\@let@token\@onedot}
\def\@onedot{\ifx\@let@token.\else.\null\fi\xspace}

\def\eg{\emph{e.g}\onedot} 
\def\ie{\emph{i.e}\onedot}

\def\etal{\emph{et al}\onedot}
\makeatother
\usepackage{colortbl}

\ifCLASSINFOpdf
\else
\fi

\begin{document}

\title{Benchmarking Micro-action Recognition: Dataset, Methods, and Applications}

%


\markboth{IEEE Transactions on Circuits and Systems for Video Technology, 2024}%
{Shell \MakeLowercase{\textit{et al.}}: Bare Demo of IEEEtran.cls for IEEE Journals}
%


\author{Dan Guo$^{\orcidlink{0000-0003-2594-254X}}$, ~\IEEEmembership{Senior Member,~IEEE}, Kun Li$^\ast$$^{\orcidlink{0000-0001-5083-2145}}$, Bin Hu$^{\orcidlink{0000-0003-3514-5413}}$,~\IEEEmembership{Fellow,~IEEE}, Yan Zhang, \\ and Meng Wang$^\ast$$^{\orcidlink{0000-0002-3094-7735}}$,~\IEEEmembership{Fellow,~IEEE}

\thanks{
This work was supported in part by the National Key Research and Development Program of China under Grant 2022YFB4500600; and in part by the National Natural Science Foundation of China under Grant 62272144, Grant 72188101, Grant 62020106007, and Grant U20A20183. 
\textit{(Corresponding authors: Kun Li; Meng Wang.)}
}

\thanks{
Dan Guo is with the Key Laboratory of Knowledge Engineering with Big Data, Ministry of Education, and School of Computer Science and Information Engineering, Hefei University of Technology (HFUT), Hefei, 230601, China, also with the Hefei Comprehensive National Science Center, Institute of Artificial Intelligence, Hefei, 230026, China, and also with Anhui Zhonghuitong Technology Co., Ltd.
(e-mail: guodan@hfut.edu.cn).
}
\thanks{
Kun Li and Yan Zhang are with School of Computer Science and Information Engineering, Hefei University of Technology (HFUT), Hefei, 230601, China. (e-mail: kunli.hfut@gmail.com; yanzhang.hfut@gmail.com). 
}
\thanks{
Bin Hu is with the Gansu Provincial Key Laboratory of Wearable Computing, School of Information Science and Engineering, Lanzhou University, Lanzhou, Gansu, 730000, China. (e-mail: bh@lzu.edu.cn).
}
\thanks{
Meng Wang is with the Key Laboratory of Knowledge Engineering with Big Data, Ministry of Education, and School of Computer Science and Information Engineering, Hefei University of Technology (HFUT), Hefei, 230601, China, and also with the Hefei Comprehensive National Science Center, Institute of Artificial Intelligence, Hefei, 230026, China. (e-mail: eric.mengwang@gmail.com).
}
}


\maketitle

\begin{abstract}
Micro-action is an imperceptible non-verbal behaviour characterised by low-intensity movement. It offers insights into the feelings and intentions of individuals and is important for human-oriented applications such as emotion recognition and psychological assessment. However, the identification, differentiation, and understanding of micro-actions pose challenges due to the imperceptible and inaccessible nature of these subtle human behaviors in everyday life. 
In this study, we innovatively collect a new micro-action dataset designated as Micro-action-52 (MA-52), and propose a benchmark named micro-action network (MANet) for micro-action recognition (MAR) task. 
Uniquely, MA-52 provides the whole-body perspective including gestures, upper- and lower-limb movements, attempting to reveal comprehensive micro-action cues. In detail, MA-52 contains 52 micro-action categories along with seven body part labels, and encompasses a full array of realistic and natural micro-actions, accounting for 205 participants and 22,422 video instances collated from the psychological interviews. 
Based on the proposed dataset, we assess MANet and other nine prevalent action recognition methods. 
MANet incorporates squeeze-and-excitation (SE) and temporal shift module (TSM) into the ResNet architecture for modeling the spatiotemporal characteristics of micro-actions. Then a joint-embedding loss is designed for semantic
matching between video and action labels; the loss is used to better distinguish between visually similar yet distinct micro-action categories.
The extended application in emotion recognition has demonstrated one of the important values of our proposed dataset and method.
In the future, further exploration of human behaviour, emotion, and psychological assessment will be conducted in depth. The dataset and source code are released at \href{https://github.com/VUT-HFUT/Micro-Action}{https://github.com/VUT-HFUT/Micro-Action}. 
\end{abstract}


%
\IEEEpeerreviewmaketitle

\section{Introduction}

\begin{figure}[t]
\centering
\includegraphics[width=1.0\linewidth]{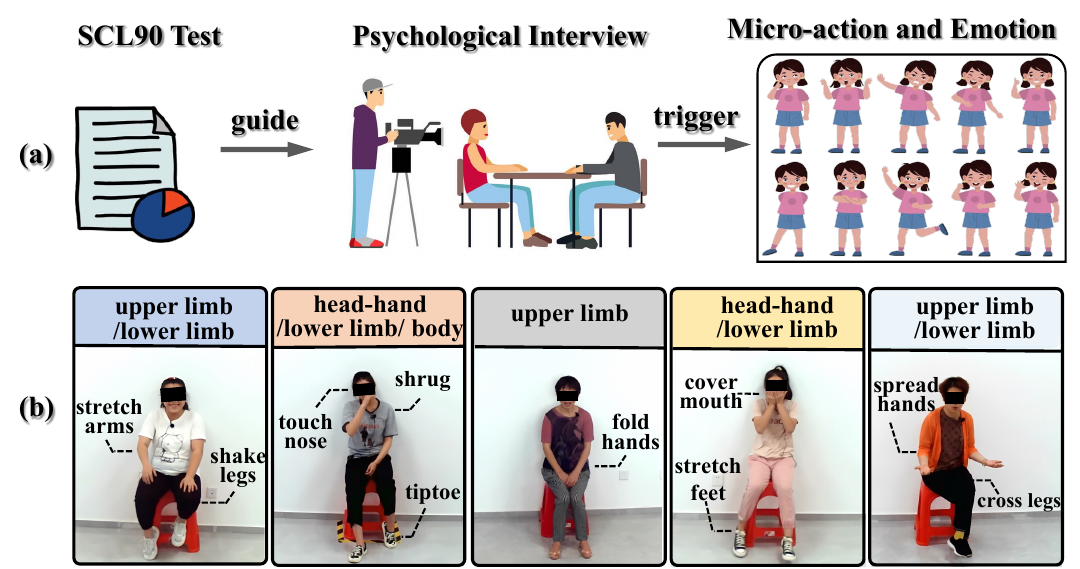}
\caption{Data collection procedure and samples of our micro-action dataset. We conduct professional face-to-face psychological interviews to collect whole-body micro-actions, focusing on two levels of prediction \ie, body part and micro-action category.  
Psychological interviews are conducted under the guidance of the Symptom Checklist 90 (SCL90) test (also called the self-report inventory)~\cite{derogatis2004scl} to elicit and collect natural spontaneous micro-actions.}
\label{fig:intro} 
\vspace{-1.2em}
\end{figure}

\IEEEPARstart{E}{nabling} the computer to intelligently detect and recognize human micro-actions has garnered increasing attention within the artificial intelligence community \cite{chen2019analyze,kopuklu2020drivermhg,li2019spatiotemporal,zhang2021detecting}. 
Researches show that micro-actions are the important human actions, generally, they can serve as the nonverbal cues to better reflect a person's mental state compared to general movements~\cite{yonetani2016recognizing,mi2019recognizing}. For example, a slight nod indicates a positive state of agreement, and a small shake in the legs may indicate a negative state such as tension. A strong Micro-action recognition (MAR) system can provide high-quality support for human-oriented technical services and innovations, it has practical implications for various fields, such as medical diagnostics~\cite{yenduri2022fine}, smart vehicles~\cite{kopuklu2020drivermhg}, face expression recognition~\cite{ge2017detecting,ge2018low}, sports competitions~\cite{shao2020finegym}, virtual reality~\cite{jourabloo2022robust}. 

Specifically, \textbf{micro-action} is characterized by rapid and subtle behaviors, exhibiting extremely low behavioral intensities. It involves whole-body movements, including the head, hands, and upper and lower limbs \cite{noroozi2018survey,guo2018hierarchical}.  
This behavior is often considered a ``silent'' form of communication, operating unconsciously and imperceptibly when individuals express their genuine thoughts, feelings, and intentions. 
Psychological theory reveals that micro-action provides a more transparent representation of individual intent than do carefully crafted verbal pronouncements~\cite{aviezer2012body}. 
Further research has also found that human behavior offers a more accurate gauge of emotion than facial expression~\cite{noroozi2018survey,xu2022emotion}. 
For instance, discerning an athlete's emotional state related to competition outcomes is challenging when relying solely on facial expressions such as weeping. However, it becomes feasible through recognition of postural behaviors~\cite{liu2021imigue}. These observations highlight the emerging effort to employ micro-actions to unveil the individuals' inner spirit or mentality.

\textbf{Micro-action recognition (MAR)} aims to detect and distinguish ephemeral body movements, generally occurring within a temporal span of 1/25s $\sim$ 1/3s. The MAR task is similar to conventional action recognition in that it uses video instances as input and requires precise and efficient algorithms. However, it is uniquely complex due to low-amplitude fluctuations in gesture and posture. 
The primary challenges of the MAR task are as follows: 
\textbf{1) Minor visual changes.} Micro-actions are subtle and rapid muscle movements that occur across various body parts, such as the head, hands, arms, legs, and feet. Identifying and distinguishing between these subtle variations in movement is a complex task.
\textbf{2) Approximate inter-class differences.} When categorizing samples from the same body part under distinct actions, visual similarity can make differentiation complicated. For example, it can be difficult to distinguish between “looking up” and “nodding” in a video clip. The former involves elevating the head, whereas the latter is characterized by repeated upward and downward head motions. 
\textbf{3) Long-tailed issue.} 
A long-tailed distribution is prevalent when the frequency of sample occurrences varies significantly across different micro-action categories.
It is difficult to avoid such data imbalance in wild data collection environments. 
For example, ``nodding'' is notably more common than ``rubbing eyes'' or ``touching ears.'' 
\textbf{4) Limited data sources.} 
The development of MAR is limited by the scarcity of participant samples and small datasets. These limitations are often worsened by the difficulty in perceiving micro-actions and concerns regarding privacy protection. 

Existing achievements in generic action recognition tasks \cite{soomro2012ucf101,kay2017kinetics,karpathy2014large,luo2021dense,wu2021spatiotemporal} focus on categorizing activities at a coarse-grained level (\eg. running and jumping), without exploring the tiny details of action variations (\eg. raising the head and nodding). Recently, there have been reports of new fine-grained action recognition tasks~\cite{liu2022fineaction,shao2020finegym,gu2020fine}. 
However, these tasks are limited to specific scenarios such as gymnastics, basketball, football, makeup, and daily life. These scenarios involve specialized behavioral patterns tailored to each action type. 

To facilitate the micro-action research, we have collected a comprehensive whole-body micro-action dataset of 52 action categories,
called Micro-Action-52 (MA-52). We recruited a significant cohort of 205 participants using a specialized face-to-face psychological interview scheme 
to capture authentic and spontaneous micro-actions for scholarly investigations. The interviewers are provided with access to the participants' SCL-90~\cite{derogatis2004scl} test results. They use open-ended questions to elicit a wide range of micro-actions from the participants. Participants are not compelled to adopt specific gestures or postures. Instead, natural expressions of their genuine spirit or mentality are encouraged. To ensure the comfort and relaxation of the participants, they remain seated during the interviews. 
As discussed in Section~\ref{sec:related}, previous works have focused on collecting upper-limb movements, whereas our dataset includes several lower-limb movements such as ``shaking legs'', ``crossing legs'', ``tiptoe'', and ``scratching feet''.

\begin{figure*}[t]
\centering
\includegraphics[width=1.0\linewidth]{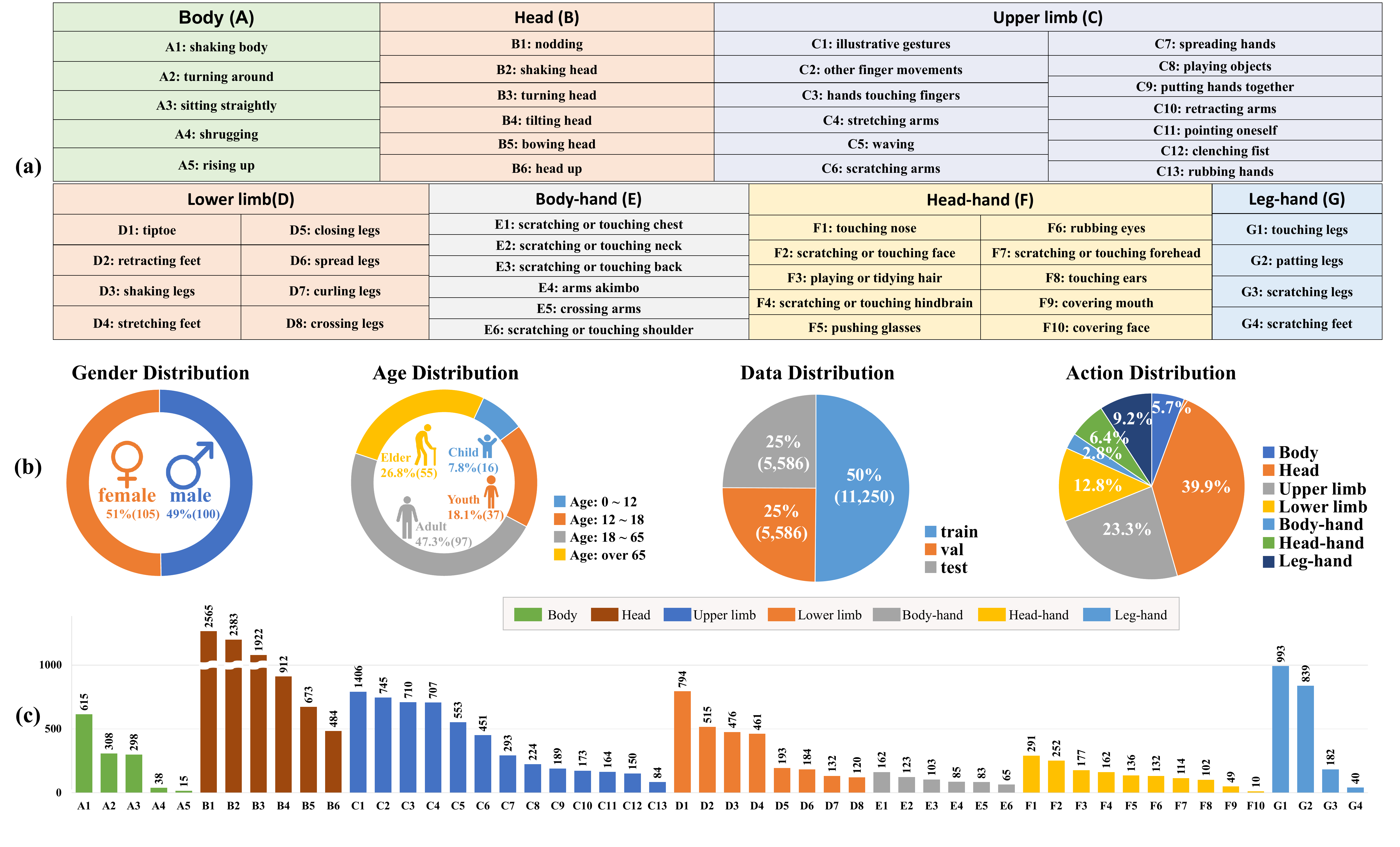}
\vspace{-1.6em}
\caption{
(a) Details of body part (coarse-grained) and micro-action (fine-grained) labels. (b) Data distribution over the gender and age of respondents, data distribution over the micro-action categories, and the proportion of training/validation/test sets. 
(c) Detailed category distribution of video samples in the MA-52 dataset over the micro-action labels.}
\vspace{-1.2em}
\label{fig:action_distribution}
\end{figure*}

The proposed MA-52 dataset comprises 52 action categories and 22,422 video samples. 
We use high-definition cameras with a resolution of 1920$\times$1080 pixels to record the interview videos.
Comparative statistics between MA-52 and the preexisting datasets are provided in Table~\ref{tab:ar_dataset}. Compared with prior studies in Table~\ref{tab:ar_dataset}, the MA-52 introduces a wealth of leg and foot movements to body language research. Furthermore, the dataset considers a greater variety of bodily interactions, involving the types of body-hand, head-hand, and leg-hand interactions, as shown in Figure~\ref{fig:action_distribution}. Overall, the MA-52 dataset stands out for its larger number of participants, extensive array of action categories, and greater volume of video instances with diverse lower-body movements and bodily interactions. These naturally occurring micro-actions can serve as reliable indicators of individual characteristics, emotional states, thought processes, and intentions. 
More details of MA-52 are elaborated in Section~\ref{sec:dataset}. 

Based on the new dataset MA-52, we propose a compatible benchmark network for micro-action recognition, namely Micro-Action Network (MANet). Furthermore, we conduct an extensive review of existing nine micro-action methods and provide a comprehensive performance comparison. To test the practicality of our fundamental data and methodology, we further investigate the application of micro-action recognition in emotion analysis.

The contributions of our work are summarized as follows: 
\begin{itemize}
\item 
A new dataset MA-52 is collected and has made publicly accessible to satisfy academic requirements for micro-action analysis. It is a comprehensive whole-body dataset consisting of a large number of video instances (22,422), a diverse participant pool (205), and arrays of body parts (7) and action categories (52). 
MA-52 is unique in its data collection method, which involves a specialized interview-based scheme. This allows participants to display natural, spontaneous, and authentic micro-actions during seated interviews, resulting in high-quality data that is rich in lower-body movements and complex bodily interactions. 
\item We present the Micro-Action Network (MANet), a benchmark network for micro-action recognition (MAR) task, and widely evaluate the suitability of prevalent general action recognition methods for MAR task. MANet outperforms existing methods, it integrates squeeze-and-excitation (SE) and temporal shift module (TSM) into the ResNet backbone and a joint-embedding loss function is designed to constrain the semantic distance between video data and action labels.
\item We further investigate the utility of our fundamental data and method in emotion analysis. We augment the dataset to create MA-52-Pro dataset and devise a dual-path MANet-based network for both micro-action and emotion recognition.  
The experiments demonstrate that capturing micro-actions significantly enhances the emotion recognition.
\end{itemize}

The remainder of this paper is organized as follows. The related works are reviewed in Section~\ref{sec:related}, and the new micro-action dataset MA-52 is introduced in Section~\ref{sec:dataset}. We elaborate on the methodologies and experiments for micro-action recognition in Sections~\ref{sec:method} and \ref{sec:experiment}. An application for emotion analysis is discussed in Section~\ref{sec:application}. Finally, we discuss future direction in Section~\ref{sec:future} and conclude this study in Section~\ref{sec:conclusion}.

\section{Related Work}
\label{sec:related}

\subsection{Tasks and Datasets} 
Human behavior analysis~\cite{guo2019dense,kuehne2011hmdb,heilbron2015activitynet,li2021proposal,fu2021learning,li2023vigt} has been extensively studied over the past few decades, with several public action datasets being released, such as 
HMDB51~\cite{kuehne2011hmdb}, ActivityNet~\cite{heilbron2015activitynet}, UCF-101~\cite{soomro2012ucf101}, Sports 1M~\cite{karpathy2014large}, and Kinetics~\cite{kay2017kinetics}. 
These datasets have contributed significantly to the development of the action recognition field.
However, they suffer from data bias~\cite{li2018resound} and do not adequately address the nuances of fine-grained activities, such as distinguishing between high jump, long jump, and triple jump in the jump type. 
The datasets contain data bias~\cite{li2018resound} due to their coarse-grained action labels, as seen in UCF-101~\cite{soomro2012ucf101} and Kinetics~\cite{kay2017kinetics}, where actions can be recognized from static frames, such as jumping and running. In this situation, it seems that action recognition no longer necessitates spatiotemporal learning. In contrast, fine-grained action recognition focuses on decomposing activities into their constituent phases and detecting minute variations between closely related actions~\cite{gu2020fine}. 
Recently, datasets such as FineGym~\cite{shao2020finegym}, Basketball~\cite{gu2020fine}, and FineAction~\cite{liu2022fineaction} have gained attention. These datasets aim to distinguish subclasses within broader action categories that typically exhibit reduced inter-class variations.
But, these are specialized behavioral patterns tailored to each action type. 

In contrast to the aforementioned efforts, micro-action recognition in the current study requires
not only the ability to differentiate fine-grained action categories but also a keen sensitivity to subtle visual alterations. 
Actions such as ``shrugging'' and ``shaking legs'' exemplify the complexity involved in capturing subtle and rapid changes in movement. The key challenges in micro-action recognition are both the identification of small movements and the differentiation between subtly distinct action categories. Thus, emerging research endeavors have been initiated to advance the state of micro-action analysis. Comparative statistics and characteristics of existing action datasets are presented in Table~\ref{tab:ar_dataset}. 
Currently, only a limited number of datasets have focused on human-centered spontaneous micro-action behaviors. For instance, the iMiGUE dataset~\cite{liu2021imigue} annotates 72 athletes across 32 behavioral classes during sports press conferences, while simultaneously recording their emotional states following the outcome of their matches. This dataset aims to interpret athletes' emotional states by analyzing their micro-gestures. 
Similarly, the Spontaneous Micro Gesture (SMG) dataset~\cite{chen2023smg} incorporates both micro-gesture and emotion recognition but requires 40 participants to narrate both fabricated and factual stories. 
The limitation here is that predetermined stories may compromise the authenticity of the participants' micro-gestures. Additionally, the Bodily Behaviors in Social Interaction (BBSI) dataset~\cite{balazia2022bodily} employs a naturalistic multi-view group conversation approach to examine the impact of body language in various social contexts, including leadership and rapport, and only captures 15 different body movements such as ``gesture,'' ``adjusting clothing,'' ``scratching.'' 
In contrast, our work utilizes a professional face-to-face psychological interview procedure and recruits an extensive cohort of 205 participants to collect genuine micro-actions. 
The dataset contains 52 distinct micro-action categories captured within the confines of a seated posture, thus facilitating the inclusion of complex leg and foot movements such as ``crossing legs'' and ``stretching feet.'' 
All the spontaneous movements provide valuable insights into individuals' cognitive processes, emotional states, and intentions. Detailed information about our dataset can be found in Section~\ref{sec:dataset}.
\begin{table*}
\renewcommand\arraystretch{1.2}
\centering
\caption{Data statistics from the Micro-Action 52 dataset and other widely used fine-grained action datasets. Sit and Stand refer to the sitting and standing postures, respectively. The standing posture is common in everyday life. The lower-limb micro-actions that remain seated contain richer leg and foot movements than those associated with standing, and the subtle and spontaneous movements captured from the lower limbs are more indicative of an individual's thoughts, emotions, and intentions.} 
\resizebox{1.0\linewidth}{!}{
\begin{tabular}{c|c|ccccc|ccc}
\hline
Datasets &Venue & Resolution & Category & Instance & Duration (s) & Participants & Source &Covered Body Area & Action type \\ \hline
ActivityNet \cite{heilbron2015activitynet} &CVPR'15 & $1280 \times 720$ &200 &23,064 &49.2 & - &Web &Whole Body & Daily events \\
HACS~\cite{zhao2019hacs} &ICCV'19 & - &200 &122,304 &33.2 & - &Web &Whole Body & Daily events  \\
FineAction~\cite{liu2022fineaction} & TIP'22 & - & 106 & 103,324 &7.1 & - & Web &Whole Body & Daily events \\ 
Diving48 \cite{li2018resound} &ECCV'18 & - & 48 & 18,404 & 5.3 & - & Web &Whole Body & Sports \\
FineGym \cite{shao2020finegym} & CVPR'20 & $1920 \times 1080$& 530 & 32,697 & 1.7 & - &Web &Whole Body & Sports \\
Basketball \cite{gu2020fine} &ICASSP'20 & - &26 & 3,399 & - & - & Web &Whole Body& Sports \\ 
MPII-cooking \cite{rohrbach2012database} &CVPR'12 & $1624 \times 1224$ & 65 & 5,609 & 11.1 m & 12 & Kitchen &Upper Body & Cooking \\
EPIC-KITCHENS \cite{damen2018scaling} &ECCV'18 & $1920 \times 1080$ &149 & 39,596 & 4.9 & 32 & Kitchen &Hands & Cooking \\
\hline
BBSI \cite{balazia2022bodily} &MM'22 &- &15 &7,905 &- & 78 & Group Conversation &Whole Body (Sit) &Social Behavior \\
\hline
iMiGUE \cite{liu2021imigue} & CVPR'21 & $1280 \times 720$ & 32 & 18,499 & 2.6 & 72 & Web &Above Chest (Sit) & Micro-gesture \\
SMG~\cite{chen2023smg} &IJCV'23 & $1920 \times 1080$ & 16 &3,712 & 1.8 & 40 & Interview &Whole Body (Stand) & Micro-gesture \\ 
MA-52 (Ours) & - & $1920 \times 1080$ & 52 & 22,422 & 1.9 & 205 &Interview &Whole Body (Sit) & Micro-action 
\\
\hline
\end{tabular}
}
\vspace{-1.2em}
\label{tab:ar_dataset}
\end{table*}

\subsection{Approaches}
Amid the rapid advancement of large-scale datasets such as ActivityNet~\cite{heilbron2015activitynet} and Kinetics~\cite{kay2017kinetics}, the field of action recognition has seen remarkable progress. Well-known methods such as  TSN~\cite{wang2018temporal}, TSM~\cite{lin2019tsm}, TIN~\cite{shao2020temporal}, C3D~\cite{C3D}, I3D~\cite{I3D}, SlowFast~\cite{feichtenhofer2019slowfast}, Video-SwinT~\cite{liu2022video}, TimeSformer~\cite{bertasius2021space}, and UniFormer~\cite{li2022uniformer} have achieved exceptionally high accuracy. However, these techniques mainly excel at recognizing generic actions such as jumping and running and deliver less accuracy in finer action categories. 
The recent focus on fine-grained action recognition has stimulated innovative approaches. Behera~\etal\cite{behera2020regional} argue that specific fine-grained actions have local discriminative semantic regions that can be evaluated using the attention mechanism. 
Consequently, they propose a regional attention network that aggregates multiple contextual regions and focuses on relevant ones. Inspired by the human visual system, Li~\etal\cite{li2022dynamic} develop a dynamic spatio-temporal specialization module that uses specialized neurons to discriminate subtle differences between fine-grained actions. Xu~\etal~\cite{xu2023pyramid} argue that regions of the body where movement occurs, such as the limbs and the trunk, provide rich semantic information. Therefore, they propose a pyramid self-attention polymerization learning framework that uses contrastive learning to jointly understand the representations of body movements and joints at different anatomical levels. 

In contrast, progress in micro-action recognition lags behind that of in generic and fine-grained action recognition due to significant challenges in dataset acquisition and labeling. 
As shown in Table~\ref{tab:ar_dataset}, the iMiGUE dataset~\cite{liu2021imigue} faces the problem of action class imbalance and long-tailed distribution, especially in uncontrolled and natural settings. 
This imbalance could cause significant performance degradation even in fully supervised learning models due to extreme label bias. To counteract this, the authors introduce an unsupervised encoder-decoder network to learn the discriminative features of micro-gestures without relying on labeled data. For the SMG dataset, Chen~\etal\cite{chen2023smg} evaluate existing skeleton and RGB-based methodologies and conclude that GCN-based skeleton models with compact network structures outperform their RGB-based counterparts. In addition, the BBSI dataset\cite{balazia2022bodily} uses four established motion feature extraction algorithms (\ie, I3D~\cite{I3D}, TSN~\cite{wang2018temporal}, TSM~\cite{lin2019tsm}, Swin Transformer~\cite{liu2021swin}) and employs the Pyramid Dilated Attention Network (PDAN)~\cite{dai2021pdan} to understand temporal relations in social behavior within group conversations. 

\subsection{Applications}
The field of micro-action recognition offers considerable possibilities and opportunities, especially in applications focused on human interaction. These techniques offer valuable insights by studying body movements, postures, and facial expressions. K{\"o}p{\"u}kl{\"u}~\etal\cite{kopuklu2020drivermhg} publish the Driver micro hand gesture (DriverMHG) dataset and a corresponding model to dynamically recognize micro hand gestures in vehicle environments. 
Gupta~\etal~\cite{gupta2023survey} describe the utility of internet of things terminals equipped with motion sensors for human-computer interaction, facilitating the capture of users' unique hand micro-movements for authentication. 
Chandio~\etal\cite{chandio2022holoset} propose the HoloSet dataset for visual-inertial odometry applications in Extended Reality (XR), capturing both macro and micro human actions. 
Recent scientific efforts have demonstrated that body movements and posture variations crucial for understanding human emotions\cite{noroozi2018survey,behera2020regional,xu2022emotion}. 
In particular, emotion analysis is a salient application of micro-action recognition~\cite{liu2021imigue,chen2023smg,deng2020mimamo,luo2020arbee,zhang2022short}. 
Numerous studies have been conducted to detect human emotional states by observing changes in facial expression, limb movement, and overall body posture. For example, Deng~\etal\cite{deng2020mimamo} capture both micro and macro facial cues for emotion prediction, while Liu~\etal\cite{liu2021imigue} use spontaneous upper limb actions, and Luo~\etal~\cite{luo2020arbee} design a holistic model that includes body components, inherent intentions, shapes, and spaces for the same purpose.

\section{New Dataset}
\label{sec:dataset}
\subsection{Data Collection and Annotation}\label{sec:motivation}
Nevertheless, the capture of spontaneous micro-actions poses significant challenges, primarily because of their rapid occurrence at low motion intensities and their inherent difficulty in detection and annotation. Owing to limited data and technological barriers, the field of micro-action recognition remains nascent. 
Current data collection can be categorized into two types. The first type involves the annotation and recording of tiny human behaviors in short film clips or TV shows~\cite{liu2021imigue,luo2020arbee}, which has the advantage of realistic scenarios, but suffers from issues such as unfavorable camera angles. 
The second approach involves inducing micro-actions in controlled laboratory settings~\cite{ranganathan2016multimodal,gavrilescu2015recognizing} using stimuli such as emotionally intense videos or instructing actors to portray specific emotions and behaviors in simulated environments~\cite{chen2019analyze,chen2023smg}. 
Although these environments can provoke a range of human behaviors, the elicited micro-actions may lack authenticity, being either exaggerated or suppressed. 
In contrast to these works, this study adopts a communicative interview approach, inviting participants to engage in a conversation with a professional psychological counselor. 

In order to optimally capture unconscious and spontaneous human micro-actions with a high degree of naturalism, certain challenges inherent to the psychological interview setting must be overcome. 
{\textbf{1) Elicitation of spontaneous micro-actions.} First, the participants are administered the SCL90-Test~\cite{derogatis2004scl}, a comprehensive 90-item psychological assessment questionnaire. 
A professional counselor then takes on the role of interview facilitator and asks a series of open-ended questions. 
These questions are designed to provoke deep introspection and detailed self-description, thereby encouraging participants to share rich information about themselves. 
To gauge participants' genuine feelings, thoughts, and intentions, the counselor provides timely feedback and asks follow-up questions, facilitating the emergence of nonverbal cues. In this study, we focus on their gestures and postures. 
{\textbf{2) Collecting whole-body micro-actions.} 
High-definition cameras are used to record unconscious behavior from a whole-body perspective during the interviews. These behaviors encompass movements of various parts of the body, including the head, hands, and both upper and lower limbs, as well as interactive movements between different parts of the body (\eg, head-hand and leg-hand interactions). 
To our knowledge, our dataset is unique in capturing participants while seated, thus providing additional data on lower-limb activities such as ``leg shaking,'' ``touching legs,'' and ``patting legs.'' 
{\textbf{3) Ensuring annotation quality.}} 
A rigorous three-step supervisory process oversees the complete data annotation. 
The first step involves self-censorship by volunteer annotators, each responsible for independently editing and annotating one complete video. This step facilitates the identification and correction of potential errors. 
The second step is cross-checking, which aims to reduce biases and inaccuracies that might arise from the volunteers' subjective interpretations of the annotations. 
When discrepancies arise, they are reconciled based on a majority consensus. In the final step, a third-party team reviews the data annotations for ultimate verification following the completion of all micro-action video annotations. 

\begin{figure*}[!t]
\centering
\includegraphics[width=1.0\linewidth]{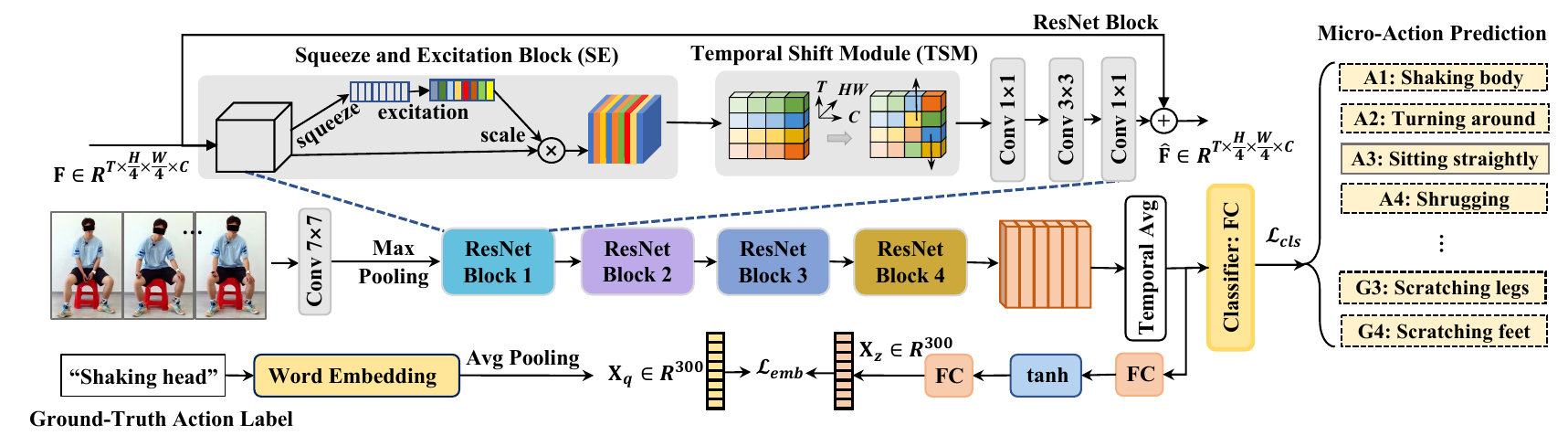}
\caption{The architecture of the Micro-action Network (MANet). The core architecture of the MANet integrates the squeeze-and-excitation (SE) \cite{hu2018squeeze} and temporal shift module (TSM)\cite{lin2019tsm} into the ResNet-50 framework. The SE specializes in channel-wise feature enhancement on the spatial feature map, whereas the TSM elevates the temporal modeling by swapping the channels of adjacent frame representations. To semantically align the action label with the video, a semantic embedding loss between the action label and the video feature is used to supervise their semantic alignment through joint feature embedding, denoted as $\mathcal{L}_{emb}$. 
The model predicts the fine-grained action label of the video under the supervision of the classification loss $\mathcal{L}_{cls}$ and the embedding loss $\mathcal{L}_{emb}$.}
\vspace{-1.2em}
\label{fig:method_ar} 
\end{figure*}

\subsection{Dataset Statistics and Properties}
\label{sec:property}
The Micro-Action-52 dataset (abbreviated as MA-52) comprises 205 participants and 22,422 video instances of micro-actions. Each video instance is recorded at a resolution of 1920$\times$1080 pixels and at a frame rate of 30 fps. 
The duration of these instances ranges from 1s to 7s, with an average duration of 1.9s, cumulating to a total span of 12.29h. The dataset is split into training, validation, and test subsets at a ratio of 2:1:1, consisting of 11,250, 5,586, and 5,586 instances, respectively. 
Before data collection, we informed each participant of the requirements for data collection and obtained their signed informed consent for academic purposes. 
We categorized micro-actions based on a two-level labeling system: coarse-grained categories included seven body parts (\ie, body, head, upper limb, lower limb, head-hand, body-hand, and leg-hand), and fine-grained categories contained 52 distinct micro-action categories. 

As shown by the statistics listed in Table~\ref{tab:ar_dataset}, the proposed dataset exhibits the following properties: 
\textbf{1) Large dataset.} 
The MA-52 dataset we propose is the largest repository of micro-actions, featuring 205 participants and 22,422 micro-action instances.
\textbf{2) Professional psychological interview.} 
To ensure the acquisition of realistic and authentic micro-actions, we invited professional counselors to host the interview. 
\textbf{3) Whole-body micro-actions.}
Compared to existing datasets, our data collection surpasses existing strategies by capturing a broader spectrum of whole-body micro-actions, with particular emphasis on lower-limb movements (\eg, crossing legs, shaking legs, and stretching feet). 
{\textbf{4) Diverse participants.}} 
Conscious efforts are made to achieve balance in gender and age demographics. Although the challenge of long-tailed distribution persists, we attempt to render the dataset as representative as feasible in an uncontrolled setting. 
It includes 16 children (7.8\%), 37 young individuals (18.1\%), 97 middle-aged participants (47.3\%), and 55 elderly individuals (26.8\%). Additionally, the MA-52 dataset comprises 100 male (49\%) and 105 female (51\%) participants. 
{\textbf{5) Two-level (coarse-to-fine) annotation.} 
The dataset is annotated in a two-level annotation system: 52 micro-action categories are aggregated under seven body labels (\ie, body, head, upper limb, lower limb, head-hand, body-hand, and leg-hand). In terms of the closest relevance, iMiGUE~\cite{liu2021imigue} offers 32 categories of micro-gestures distributed across five groups (\ie, body, head, hand, body--hand, and head--hand). Our annotations strive for granularity in distinguishing similar yet distinct micro-actions. For instance, in MA-52, ``nodding head'' and ``bowing head'' are categorized under the coarse-grained ``head'' label. 
\textbf{6) High-resolution data.} As the statistics shown in Table~\ref{tab:ar_dataset}, our dataset portrays the instances of high-resolution 1920$\times$1080 video. 
Thus, compared with existing datasets, our dataset delivers significant advantages in terms of action categories, action instances, participants, and video format.

\section{Methodologies}
\label{sec:method}
Owing to the inherent complexity of micro-actions, distinguishing
between similar yet distinct categories of micro-actions remains an arduous challenge. To address this task, this section introduces a methodology established on the ResNet backbone~\cite{he2016deep} combined with squeeze-and-excitation (SE)~\cite{hu2018squeeze} and temporal shift module (TSM)~\cite{lin2019tsm} for micro-action recognition and also examines existing generic action recognition algorithms. 

\subsection{Task Definition} 
Consider a human micro-action video defined as $\mathcal{V}$ = $\{{I}_1,{I}_2,\dots,{I}_T\}$, where $T$ is the length of the video. 
Our goal is to classify the micro-actions contained in the video by selecting them from a set of micro-action labels $\mathcal{Y}$. 
We annotate the affiliation relationship between (fine-grained) micro-actions $\{y_1,...,y_{N^F_A}\}$ and (coarse-grained) body parts $\{y_1,...,y_{N^C_A}\}$, where ${N^F_A}$ and ${N^C_A}$ represent the number of micro-action categories and body parts, respectively. The recognized fine-grained micro-action category serves as an indicator of its corresponding coarse-grained body part. 

\subsection{General Action Recognition Benchmarks} 
In this work, we evaluate nine existing algorithms specifically designed for generic action recognition tasks. These algorithms fall into three different categories: 1) {2D CNN-based methods} (TSN~\cite{wang2018temporal}, TSM~\cite{lin2019tsm}, and TIN \cite{shao2020temporal}); 2) {3D CNN-based methods} (C3D~\cite{C3D}, I3D~\cite{I3D} and SlowFast~\cite{feichtenhofer2019slowfast}), and 3) {Transformer-based methods} (VSwinT~\cite{liu2022video}, TimeSF~\cite{bertasius2021space}, and UniF~\cite{li2022uniformer}). For implementation, we retain the architecture of these algorithms and append a fully connected (FC) layer to act as an action classifier. The classifier produces a probability vector for all micro-action categories, written as $\mathbf{y}=\mathbf{W} \cdot \mathbf{X}+b$, where $\mathbf{X}$ is the representation vector of the video obtained from the benchmark algorithms, $\mathbf{W}$ is a learnable parameter, and $b$ is the bias. 

\subsection{Our Method} 
\textbf{Pipeline Overview.} 
The pipeline of our method is illustrated in Figure~\ref{fig:method_ar}. We adopt the ResNet architecture as the backbone and incorporate the Temporal Squeeze-and-Excitation (SE)~\cite{hu2018squeeze} and Temporal Shift Module (TSM)~\cite{lin2019tsm} and into it for micro-action recognition. 
We refer to this composite model as MANet, which offers the advantages of both SE and TSM: SE extends the spatial feature map via channel-wise excitation, while TSM facilitates temporal information exchange by reallocating a subset of channels along the timeline. 
Thus, each module uniquely addresses either spatial or temporal learning. 
First, we map the original video $\mathcal{V}$ into a feature map $\mathbf{F}\in \mathbb{R}^{T\times H/4\times W/4\times C}$ through a convolutional layer with a kernel size of $7\times7$ and a max pooling layer. 
Then a new ResNet block is equipped with SE~\cite{hu2018squeeze} and TSM~\cite{lin2019tsm}. $\mathbf{F}$ is fed into a series of ResNet blocks for spatio-temporal modeling. 
For each block, the first step is SE-$squeeze$, which uses an average pooling to compress $\mathbf{F}\in \mathbb{R}^{T\times H/4 \times W/4\times C}$ into $\mathbf{z} \in \mathbb{R}^{T\times1\times1\times C}$. The second step involves SE-$excitation$, which consists of two FC layers. 
The first FC layer contains $ \frac{C}{Ratio}$ neurons and maps $\mathbf{z}$ into the dimension $\mathbb{R}^{T\times1\times1\times\frac{C}{Ratio}}$; the second FC layer contains $C$ neurons and reshapes the $\mathbf{z}$ back to the original dimension $\mathbb{R}^{T\times1\times1\times C}$. We set Ratio=4. 
The three steps include performing the element-wise multiplication is performed on the channel weight with the original feature map $\mathbf{F}\in \mathbb{R}^{T\times H/4 \times W/4\times C}$; the new feature $\mathbf{F}^{\prime}$ is obtained. 
Furthermore, we conduct the TSM operation. We split $\mathbf{F}^{\prime}$ into eight chunks and shift the chunks along the temporal dimension. 
The process is formulated as follows:
\begin{eqnarray}
\begin{aligned}
& \hat{\mathbf{F}} = \rm{TSM}(\mathbf{F}^{\prime}) \in \mathbb{R}^{T\times H/4\times W/4\times C}, \\
&\Leftrightarrow \left\{\begin{array}{l}
\mathbf{F}^{\prime} = [\mathbf{f}_1^{\prime}, \cdots, \mathbf{f}_T^{\prime}] \in \mathbb{R}^{T\times H/4\times W/4\times C};\\
\mathbf{f}_t^{\prime\prime} \leftarrow (\mathbf{f}^{\prime}_{t-1},\mathbf{f}^{\prime}_t,\mathbf{f}^{\prime}_{t+1}; 1/8);\\
\hat{\mathbf{F}} = ReLu(Conv(\mathbf{F}^{\prime\prime})),
\end{array}\right.
\end{aligned}
\label{eq:tsm}
\end{eqnarray}
where 1/8 denotes the shifted length and the kernel size of the convolutional layer $Conv$ is set to $3 \times 3$. 
In the last step of the ResNet block, the feature $\hat{\mathbf{F}}$ is fed to three convolution layers with the kernel sizes of $1\times1$, $3\times3$, and $1\times1$, and then summed up with $\mathbf{F}$. We implement four ResNet blocks and output the video feature $\mathbf{X}\in \mathbb{R}^{T\times H/32 \times W/32 \times 32C}$. 

\textbf{Embedding-based Optimization.} 
The basic objective function employed for this task is the cross-entropy loss, denoted as $\mathcal{L}_{cls}$. This loss constrains the predicted probability corresponding to the correct action. Additionally, an embedding loss $\mathcal{L}_{emb}$ is incorporated to align the semantic distance between the video and its ground-truth action label. 
Specifically, the action labels are transformed into GloVe embedding vectors~\cite{pennington2014glove}. 
The embedding vector of each action label, denoted as $\mathbf{X}_{q} \in \mathbb{R}^{300}$, is obtained via the average pooling of the action words for each video instance. 
The averaged visual features of the video are then mapped into a joint embedding space, represented as $\mathbf{X}_{z} \in \mathbb{R}^{300}$. 
A semantic constraint is imposed by computing the Euclidean distance between $\mathbf{X}_{q}$ and $\mathbf{X}_{z}$. The embedding loss formulation is given as follows: 
\begin{equation}
\mathcal{L}_{emb} = {\Vert \mathbf{X}_{q} - \mathbf{X}_{z} \Vert}^2.
\end{equation} 

The total loss function is formulated as follows: 
\begin{equation}
\mathcal{L} = \mathcal{L}_{cls} + \alpha \mathcal{L}_{emb},
\label{eq:loss}
\end{equation}
where $\alpha$ denotes a balance hyperparameter for the loss terms.

\begin{table*}[!th]
\renewcommand\arraystretch{1.3}
\centering
\caption{Performance comparison for micro-action recognition on the Micro-action 52 dataset. The best are highlighted in \textbf{bold}.}
\resizebox{1.0\linewidth}{!}{
\begin{tabular}{c|c|ccc|ccc|ccc|c}
\hline
Action Label & Metric & TSN\cite{wang2018temporal} & TSM\cite{lin2019tsm} &TIN\cite{shao2020temporal} & C3D\cite{C3D} & I3D\cite{I3D} & SlowFast\cite{feichtenhofer2019slowfast} &VSwinT\cite{liu2022video} &TimeSF\cite{bertasius2021space} &UniF\cite{li2022uniformer} &\textbf{MANet (Ours)} \\ \hline
$fine, coarse$ & F1$_{\emph{mean}}$ &43.67 & 61.39 & 58.22 & 58.43 & 61.66 & 63.09 & 61.24 & 51.53 & \underline{64.43 }& \textbf{65.59} \\
\hline
$coarse$ &F1$_\emph{{macro}}$ &52.50 & 70.98 & 66.99 & 66.60 & 71.56 & 70.61 & 71.25 & 61.90 & \underline{71.80} & \textbf{72.87} \\
$fine$ &F1$_\emph{{macro}}$ & 28.52 & 40.19 & 39.82 & 40.86 & 39.84 & 44.96 & 38.53 & 34.38 & \underline{48.01} & \textbf{49.22}\\
\hline
$coarse$ &F1$_\emph{{micro}}$ & 59.22 & 77.64 & 73.26 & 74.04 & 78.16 & 77.18 & 77.95 & 69.17 & \textbf{79.03} &\underline{78.95} \\
$fine$ &F1$_\emph{{micro}}$ &34.46 & 56.75 & 52.81 & 52.22 & 57.07 & \underline{59.60}  & 57.23 & 40.67 & {58.89} & \textbf{61.33} \\
\hline
$coarse$ &Acc-Top1 & 59.22  &77.64 & 73.26 & 74.04 & 78.16 & 77.18 & {77.95} & 69.17 & \textbf{79.03} & \underline{78.95} \\
$fine$ &Acc-Top1 & 34.46 & 56.75 & 52.81 & 52.22 & 57.07 & \underline{59.60} &{57.23} & 40.67 & 58.89 & \textbf{61.33} \\
$fine$ &Acc-Top5 &73.34 & 87.47 & 85.37 & 86.97 & \underline{88.67} & 88.54 & 87.99  & 82.62 & 87.29 & \textbf{88.83} \\ \hline
\end{tabular}}
\label{tab:act-cmp}
\vspace{-1.2em}
\end{table*}

\begin{table*}[t!]
\renewcommand\arraystretch{1.3}
\centering
\caption{Ablation results of sampling $T$ frames per video on the Micro-action 52 dataset. The best are highlighted with \textbf{bold}.}
\vspace{0.5em}
\resizebox{1.0\linewidth}{!}{
\begin{tabular}{c|cccccccc}
\hline
Metric & F1$_{\emph{mean}}$ (fine, coarse) & F1$_{\emph{macro}}$ (coarse) & F1$_{\emph{macro}}$ (fine) & F1$_{\emph{micro}}$ (coarse) & F1$_{\emph{micro}}$ (fine) & Acc-Top1 (coarse) & Acc-Top1 (fine) & Acc-Top5 (fine) \\ \hline
$T$=4  &61.20 & 69.22 &45.72 &75.30 &54.57 &75.30 &54.57 &87.54 \\
$T$=\textbf{8}  & \textbf{65.59} & \textbf{72.87} & \textbf{49.22} & \textbf{78.95} & \textbf{61.33} & \textbf{78.95} & \textbf{61.33} & {88.83} \\
$T$=16 & 64.66 & 71.77 & 47.78 & 78.32 & 60.76 & 78.32 & 60.76 & \textbf{88.95} \\ \hline  
\end{tabular}}
\vspace{-1.2em}
\label{tab:abl_t}
\end{table*}

\begin{table}[t!]
\renewcommand\arraystretch{1.34}
\centering
\caption{Ablation results of hyperparameter $\alpha$ in loss function for micro-action recognition. The best results are highlighted in \textbf{bold}.}
\resizebox{1.0\linewidth}{!}{
\begin{tabular}{c|c|cccccc}
\hline
Action Label &Metric & $\alpha$=0 & $\alpha$=0.1 & $\alpha$=1 & $\alpha$=2 & $\alpha$=5 & $\alpha$=10 \\ \hline
$fine, coarse$ &F1$_{\emph{mean}}$ & 63.55 & 63.65 & 63.77 & 62.51 & \textbf{65.59} & 64.11 \\
$coarse$ &F1$_\emph{{macro}}$ & 70.76 & 70.70 & 71.02 & 71.73 & \textbf{72.87} & 72.18\\
$fine$ &F1$_\emph{{macro}}$ & 47.97 & {48.38} & 48.10 & 47.56 & \textbf{49.22} & 47.16 \\ \hline
$coarse$ &F1$_\emph{{micro}}$ & 76.94 & 77.16 & 77.72 & 78.12 &\textbf{78.95} & 78.31 \\
$fine$ &F1$_\emph{{micro}}$ & 58.54 &58.80 & 58.50 &58.96 &\textbf{61.33} & 59.71\\ \hline
$coarse$ &Acc-Top1 & 76.94 & 77.17 & 77.72 & 78.12 &\textbf{78.95} & 78.31\\ 
$fine$ &Acc-Top1 & 58.54 &58.80 & 58.50 & 58.96 &\textbf{61.33} & 59.71\\
$fine$ &Acc-Top5 & \textbf{89.12} & 88.36 & 88.47 & 88.31 & {88.83} & 86.74 \\ 
\hline
\end{tabular}
}
\vspace{-1.2em}
\label{tab:actloss-abl}
\end{table}

\section{Experiments}
\label{sec:experiment}
\subsection{Evaluation Metrics}
\label{metrics}
We adopt standard evaluation metrics for generic action recognition tasks, including Accuracy (Acc) and F1 score~\cite{bao2021evidential}. Specifically, we compute Acc-Top1 and Acc-Top5 under the categories of ``coarse'' and ``fine'' action labels. 
The F1 metric is particularly relevant in scenarios with unbalanced data and is implemented in variants such as F1$_\emph{{micro}}$ and F1$_\emph{{macro}}$. 
Given the long-tailed distribution of micro-action categories in our data, all F1 metrics are used to provide comprehensive evaluations from multiple perspectives. 
Importantly, the terms ``micro'' and ``macro'' in F1$_\emph{{micro}}$ and F1$_\emph{{macro}}$ refer solely to their metric calculation methods and should not be confused with ``fine-grained'' and ``coarse-grained'' labels in our dataset.

In detail, F1$_\emph{{micro}}$ evaluates the overall true positives, false negatives, and false positives, while F1$_\emph{{macro}}$ calculates the unweighted mean for each action category. F1$_\emph{{micro}}$ treats all samples equally and is less influenced by any category with a predominant number of micro-actions. Conversely, F1$_\emph{{macro}}$ weights the contributions of each category equally. The metrics are expressed as follows: 
\begin{equation}
\rm F1_\emph{{micro}}\!=\!2\cdot \frac{\bar{Pre} \!\cdot\! \bar{Recall}}{\bar{Pre}\!+\!\bar{Recall}}, 
\textrm{s.t.}\!\left\{\begin{array}{l}
\! \rm \bar{Pre} \!=\! \frac{\sum_{\emph{i}=1}^\emph{N} TP_\emph{{i}}}{\sum_{\emph{i}=1}^\emph{N} TP_\emph{{i}} + \sum_{\emph{i}=1}^\emph{N} {FP}_\emph{{i}}};\\
\! \rm \bar{Recall} \!=\! \frac{\sum_{\emph{i}=1}^\emph{N} TP_\emph{{i}}}{\sum_{\emph{i}=1}^\emph{N} TP_\emph{{i}} + \sum_{\emph{i}=1}^\emph{N} FN_\emph{{i}}},\\
\end{array}\right.
\end{equation}
\begin{equation}
\label{eq:macro}
\rm F1_\emph{{macro}}\!=\!2\cdot \frac{\dot{Pre} \cdot \dot{Recall}}{\dot{Pre} +\dot{Recall}},
\textrm{s.t.}\!\left\{\begin{array}{l}
\rm \dot{Pre} \!=\! \sum_{\emph{j}=1}^{\emph{N}_\emph{C}} Pre_\emph{{j}}/{\emph{N}_\emph{C}} ;\\
\rm \dot{Recall} \!=\! \sum_{\emph{j}=1}^{\emph{N}_\emph{C}} Recall_\emph{{j}}/{\emph{N}_\emph{C}},\\
\end{array}\right.
\end{equation}
where $N$ is the number of samples and $N_\emph{C}$ denotes the number of micro-action categories; Pre$_\emph{j}$ denotes the predicted precision for micro-action category $\emph{j}$. \{TP$_\emph{i}\}$, \{FN$_\emph{i}\}$, and \{FP$_\emph{i}\}$ are the true positives, false negatives, and false positives in the dataset.

With regard to the two-level (coarse- and fine-grained) action labels in the present dataset, we denote the metrics as F1$_{\emph{macro}}^{\emph{coarse}}$, F1$_{\emph{macro}}^{\emph{fine}}$, F1$_{\emph{micro}}^{\emph{coarse}}$, and F1$_{\emph{micro}}^{\emph{fine}}$. 
We calculate a mean F1 as follows: 
\begin{equation}
\rm F1_{\emph{mean}} =\frac{\rm F1_{\emph{macro}}^{\emph{coarse}} \!+\! \rm F1_{\emph{micro}}^{\emph{coarse}} \!+\! \rm F1_{\emph{macro}}^{\emph{fine}} \!+\! \rm F1_{\emph{micro}}^{\emph{fine}} }{4}.
\end{equation}

\subsection{Implementation Details} In the MA-52 dataset, the average video duration is 1.9s. Therefore, we empirically sample $T=8$ frames from each video sample and resize each frame to $224\times224$.
The MANet is constructed on the ResNet~\cite{he2016deep} and equipped with SE \cite{hu2018squeeze} and TSM \cite{lin2019tsm}. 
The SE ratio is set to 4, while the temporal shift length in TSM is set to $1/8$. For model optimization, the balancing hyperparameter $\alpha$ in Eq.~\ref{eq:loss} is set to 5.  
We set the SGD optimizer with a learning rate of 0.001, a momentum of 0.9, a weight decay of 1e-4, and a batch size of 10 for model training. 
The learning rate is reduced by a factor of 10 at the 30th and 60th epochs, and the model is trained with 80 epochs. 
All experiments are run on the PyTorch platform.

\subsection{Experimental Analysis}
\textbf{Main Comparison.} 
As shown in Table~\ref{tab:act-cmp}, in the context of 2D CNN, 3D CNN, and Transformer models, TSM~\cite{lin2019tsm}, SlowFast~\cite{feichtenhofer2019slowfast}, and UniFormer~\cite{li2022uniformer} achieve commendable performance, securing F1$_{\textit{mean}}$ scores of 61.39\%, 63.09\%, and 64.43\%, respectively. 
In particular, UniFormer stands out for its exceptional feature representation and global context modeling. 
Our proposed MANet outperforms UniFormer by 1.16\% on F1$_{\textit{mean}}$, thus demonstrating the effectiveness of our methodology. 
Observing the experimental results, regardless of coarse- or fine-grained labels, our model sets new benchmarks on all major evaluation metrics such as F1$_{\textit{micro}}$, F1$_{\textit{macro}}$, F1$_{\textit{mean}}$, Acc-Top1, and Acc-Top5. 
For example, MANet outperforms UniFormer by 2.44\% on F1$_{\textit{micro}}^{\textit{fine}}$ and outperforms the baseline TSM~\cite{lin2019tsm} by 4.20\% on F1$_{\textit{mean}}$. 

Furthermore, we observe that fine-grained micro-action recognition is inherently more challenging than coarse-grained recognition. As shown in Table~\ref{tab:act-cmp} again, the results of fine-grained evaluations are consistently lower than their coarse-grained counterparts across all methods. Specifically, MANet's F1$_{\textit{macro}}$ score is 72.87\% in the coarse-grained setting, but drops sharply to 49.22\% in the fine-grained setting. 
This trend is also evident in the Acc-Top1 metric, where the coarse-grained body part Acc-Top1 exceeds 78.95\% but falls below 61.33\% in the fine-grained micro-action recognition. This phenomenon is consistent with fine-grained generic action recognition~\cite{li2022dynamic,xu2023pyramid,behera2020regional,liu2021imigue,gao2021skeleton}, which is characterized by a wide range of action categories.

\textbf{Ablation Studies.}
We perform ablation experiments of main components, loss hyperparameter $\alpha$, and the number of sampling frames per video on all of the evaluation metrics. The experimental results demonstrate the stability of our model. There are three variants of MANet: 1) \textbf{baseline}: TSM \cite{lin2019tsm}, 2) \textbf{w/o SE}: removing SE \cite{hu2018squeeze} from MANet, and 3) \textbf{w/o emb}: removing the label embedding loss from MANet. 
From Table \ref{tab:act-abl}, we can see that adding either \textbf{SE} or \textbf{emb} into the model benefits the performance improvement over the  \textbf{baseline}. 
\textbf{w/o emb} and \textbf{w/o SE} perform 2.04\% and 0.82\% lower than MANet on the F1$_{\emph{mean}}$ metric in the setting of \{fine, coarse\} labels, respectively. Especially for \textbf{w/o emb}, the performance drop demonstrates the effectiveness of the label embedding associated with the video features. Allowing the model to distinguish the similarities between video features and micro-action labels can increase the model's ability to recognize similar but distinct micro-actions, thereby improving the model's generalization ability and accuracy. 

The hyperparameter $\alpha$ is used to balance $\mathcal{L}_{cls}$ and $\mathcal{L}_{emb}$. Its effect is to prevent $\mathcal{L}_{cls}$ from being overly dominant or $\mathcal{L}_{emb}$ from being neglected. It can effectively prevent overfitting training resulting in poor performance. In other terms, $\mathcal{L}_{emb}$ is used to strengthen the constraint at the semantic level. 
Table \ref{tab:actloss-abl} shows the results for the parameters $\alpha \in \{0,0.1,1,2,5,10\}$. Observing Table \ref{tab:actloss-abl}, $\alpha$=$0$ indicates that removing \textbf{emb} from MANet (\textbf{w/o emb}) produces the lowest performance for all the metrics. 
When $\alpha$=5, the result of F1$_{\emph{mean}}$ with \{fine, coarse\} labels is optimal at 65.59\%. The hyperparameter $\alpha$ positively affects the balance between $\mathcal{L}_{cls}$ and $\mathcal{L}_{emb}$.

\begin{table}[t!]
\renewcommand\arraystretch{1.2}
\centering
\caption{Ablation results of different MANet components on the Micro-action 52 dataset. The best results are highlighted in \textbf{bold}.}
\resizebox{1.0\linewidth}{!}{
\begin{tabular}{c|c|cccc}
\hline
Action Label &Metric & Baseline & w/o SE & w/o emb & MANet \\ \hline
$fine, coarse$ & F1$_{\emph{mean}}$ & 61.39  & 64.77 & 63.55 & \textbf{65.59} \\ \hline
$coarse$ &F1$_\emph{{macro}}$ & 70.98 & 72.22 & 70.76 & \textbf{72.87} \\
$fine$ &F1$_\emph{{macro}}$ & 40.19 & 48.07 &47.97 & \textbf{49.22} \\ \hline
$coarse$ &F1$_\emph{{micro}}$ & 77.64 & 78.48  &76.94 & \textbf{78.95} \\ 
$fine$ &F1$_\emph{{micro}}$ & 56.75 &60.31 &58.54 &\textbf{61.33} \\ \hline
$coarse$ &Acc-Top1 & 77.64 & 78.48  & 76.94 & \textbf{78.95} \\
$fine$ &Acc-Top1 & 56.75 & 60.31 & 58.54 &\textbf{61.33} \\
$fine$ &Acc-Top5 & 87.47 & {89.01} & \textbf{89.12} & 88.83 \\ \hline
\end{tabular}
}
\label{tab:act-abl}
\vspace{-1.2em}
\end{table}

We conduct the ablation experiment of the setting of $T$ in Table~\ref{tab:abl_t}. When $T$ is set to 4, the Acc-Top1 metric with fine labels drops 6.76\% compared to the setup of $T$ = 8. We speculate that important information is lost when the input frames are too small, resulting in the model being unable to capture the continuous details of micro-actions. When $T$ is set to 16, the performance is dropped by a large margin, \eg, F1$_{\textit{mean}}$ is dropped from 65.59 to 64.66. {This may be due to the fact that redundant frames interfere with micro-action acquisition.} Based on the above observations, we set the number of sampling frames to $T$ = 8, which can reduce computing costs while retaining important information to achieve the best micro-action recognition effect.

\begin{figure*}[!t]
	\centering
	\includegraphics[width=1.0\linewidth]{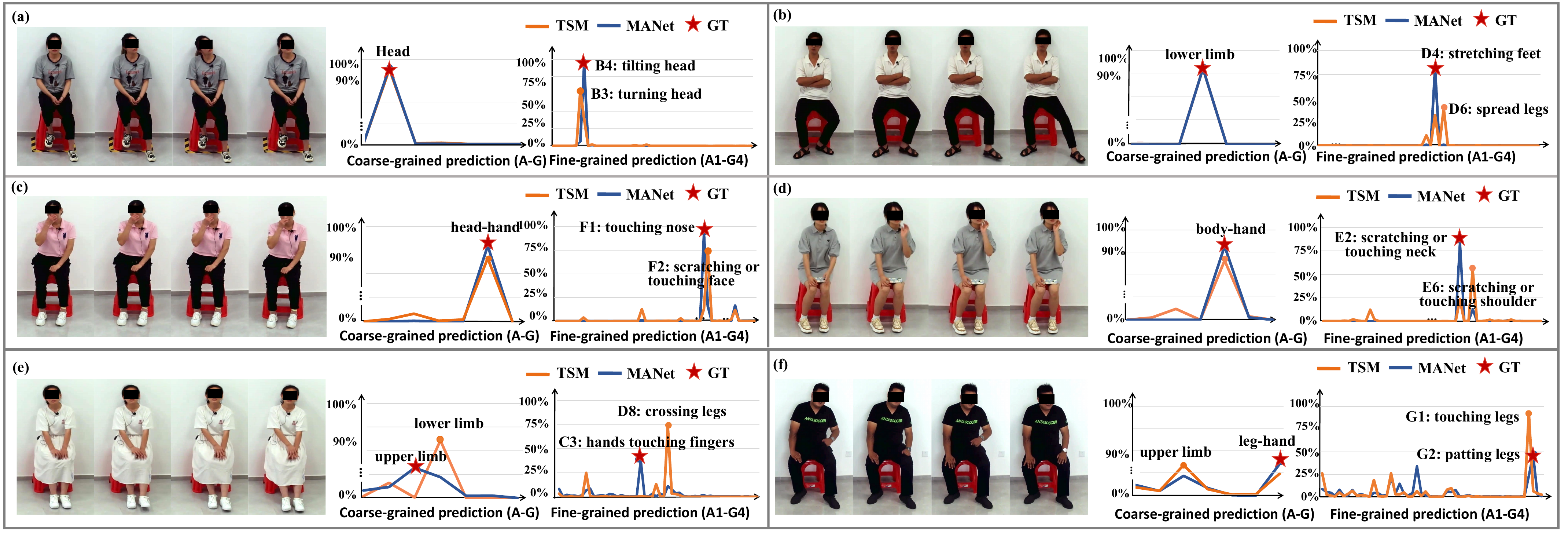}
	\caption{The prediction results of six examples for micro-action recognition examples on the MA-52 dataset are displayed, and the line graphs show the probability distributions with both coarse- and fine-grained labels. 
		The MANet model demonstrates robust performance at both coarse- and fine-grained levels. 
		For instance, Figure~\ref{fig:action_visual2} (e) reveals that MANet accurately predicts the micro-actions ``hands touching fingers'' in conjunction with the body part interaction of ``upper limb.'' Conversely, the TSM model misclassifies the coarse-grained label as ``lower limb'' and incorrectly identifies the fine-grained label as ``crossing legs'' in this specific case. The task of distinguishing between highly similar micro-actions remains a pressing challenge.}
	\vspace{-1.2em}
	\label{fig:action_visual2}
\end{figure*}

\begin{figure}[t!]
	\centering
	\includegraphics[width=1.0\linewidth]{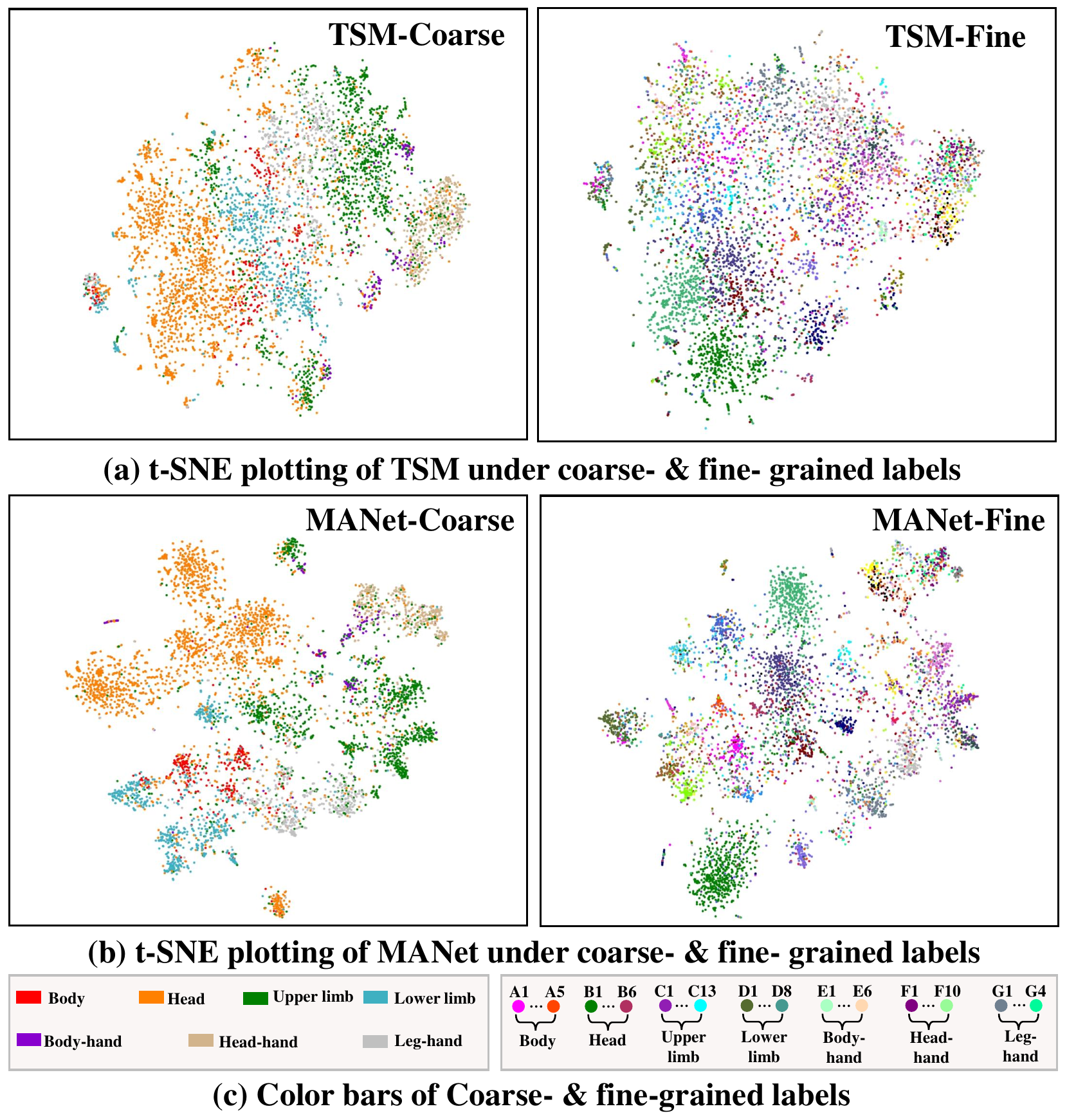}
	\caption{t-SNE~\cite{van2008visualizing} results of coarse- and fine-grained features on the test set of Micro-action-52 dataset. Each point indicates a video instance and different colors indicate various micro-action categories. Compared with TSM \cite{lin2019tsm}, MANet shows a clear clustering effect on both coarse- and fine-grained micro-action categories.}
	\vspace{-1.2em}
	\label{fig:tsne} 
\end{figure}

\begin{figure}[t!]
\centering
\includegraphics[width=1.0\linewidth]{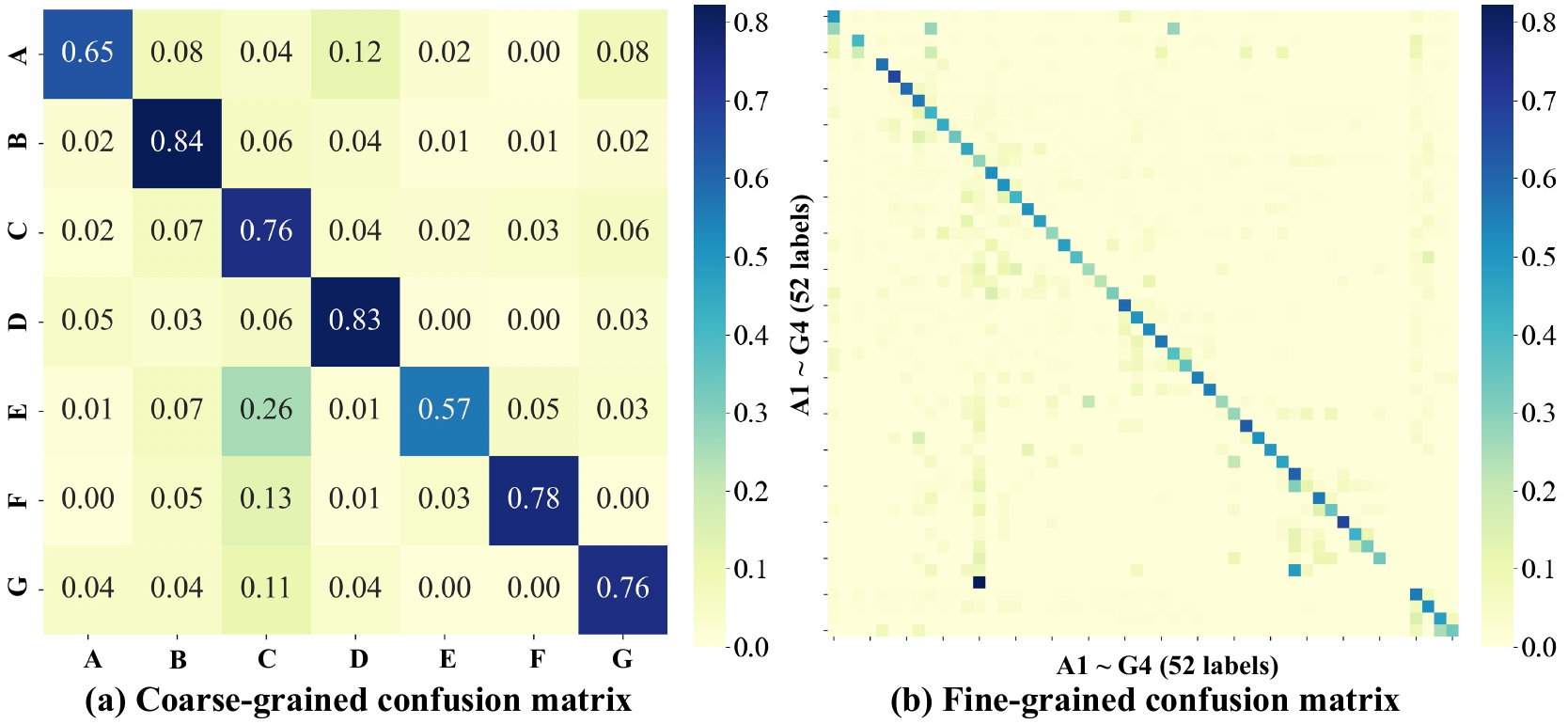}
\vspace{-1.0em}
\caption{Predicted confusion matrices of fine- and coarse-grained categories of micro-action categories from the MA-52 test set. Almost all intensive results are distributed along the diagonal. MANet provides promising prediction results that are consistent with the category-specific ground truth.}
\label{fig:action_matrix}
\vspace{-1.2em}
\end{figure}

\subsection{Visualization Examples}
In this subsection, we present visualizations from different perspectives, including feature distribution, confusion matrix, and prediction results, all focused on micro-action recognition.

\textbf{Feature Distributions.} 
We plot the feature distribution using t-SNE~\cite{van2008visualizing}. 
Specifically, the features are extracted from the layer preceding the classifier in both MANet and TSM architectures. Figure~\ref{fig:tsne} shows these features categorized into both coarse- and fine-grained labels. Notably, MANet exhibits superior clustering efficiency in distinguishing both coarse- and fine-grained categories as compared to TSM. In Figure~\ref{fig:tsne} (b), samples with micro-action labels ``B1'' and ``B6'' (``nodding'' and ``head up'') are cleanly segregated into two distinct clusters using our method. Conversely, Figure~\ref{fig:tsne} (a) presents a more amalgamated distribution of micro-action samples when using TSM. 
This evidence confirms that MANet can effectively discriminate between micro-actions that are similar but different. 

\textbf{Confusion Matrices of the Prediction Results.} 
Figure~\ref{fig:action_matrix} presents the confusion matrices for the predictions of MANet across both coarse- and fine-grained labels. 
From Figures~\ref{fig:action_matrix} (a) and (b), the relationships among the micro-actions are predominantly align along the diagonal, indicating that MANet's predictions are largely congruent with the respective category-specific ground truth. 
In addition, some micro-actions (\eg, A2: ``turning around,'' and A4: ``shrugging'') are misclassified, possibly due to approximate inter-class difference and insufficient training data. The long-tailed distribution of micro-action categories remains an issue to be resolved in this field.

\textbf{Visualization Examples.} 
Figure~\ref{fig:action_visual2} presents six comparative examples between TSM and MANet on coarse- and fine-grained recognition. 
Figures~\ref{fig:action_visual2} (a) and (b) highlight the importance of capturing the temporal dynamics of micro-actions. 
In particular, the distinction between ``tilting head'' and ``turning head'' requires the frequency of the head movement to be determined, as shown in Figure~\ref{fig:action_visual2} (a), and the distinction between ``stretching feet" and ``spread legs'' lies in the parts of the movements as shown in Figure~\ref{fig:action_visual2} (b). 
In Figure~\ref{fig:action_visual2} (c), MANet accurately identifies the micro-action as ``touching nose'', in contrast to TSM, which incorrectly identifies it as ``scratching or touching face''. 
MANet predicts the correct micro-actions. Figure~\ref{fig:action_visual2} (d) demonstrates that MANet provides superior results on the correct body part and the
correct micro-action category. 
TSM misclassifies the label ``scratching or touching neck'' and incorrectly recognizes the micro-action ``scratching or touching shoulder'' in this instance. 
Figure~\ref{fig:action_visual2} (e) describes MANet's ability to recognize subtle micro-actions such as ``hands touching fingers''.
Figure~\ref{fig:action_visual2} (f) illustrates the importance of temporal information in micro-action recognition. Our model can predict the action as ``patting legs", while TSM incorrectly predicts the action as ``touching legs". 
Distinguishing the extremely small differences between the similar actions remains an urgent challenge for this task.

\section{Application in Emotion Analysis}
\label{sec:application}
Emotion analysis is a salient direction in human-centered applications. 
Existing methodologies are mainly focus on facial expression recognition \cite{sanchez2021affective,lee2019context,zhang2022short,li2021dbcface}. 
However, these approaches are often hindered by various limiting factors such as low resolution, suboptimal illumination, restricted viewing angles, and even privacy protection. 
Recent efforts in micro-action recognition \cite{luo2020arbee,liu2021imigue,chen2023smg} have demonstrated that precise capture and interpretation of micro-action priors can offer advantageous insights for emotion analysis. 
Despite these advances, the achievements in micro-action recognition remain remarkably limited and lag significantly behind developments in psychological theories. To assess the practicality of the new data source and methodology introduced in this study, we apply them to the emotion analysis application.

\subsection{Data Preparation and Methodology} 
We extend the MA-52-Pro dataset specifically for emotion analysis, where each instance is sequentially annotated with an emotion label followed by a micro-action label. 
In this study, we collect two micro-action datasets, namely the Micro-action 52 dataset (MA-52) for fine-grained and coarse-grained micro-action recognition, and extend an MA-52-Pro dataset for multi-label micro-action recognition and emotion recognition.
Our observation shows that shifts in human emotion are frequently accompanied by the simultaneous occurrence of multiple micro-actions. Visualization samples of MA-52-Pro dataset for multi-label micro-action recognition and emotion recognition can be obsevered in Figure~\ref{fig:emo-vis2}.

\begin{figure}[!t]
\centering
\includegraphics[width=1.0\linewidth]{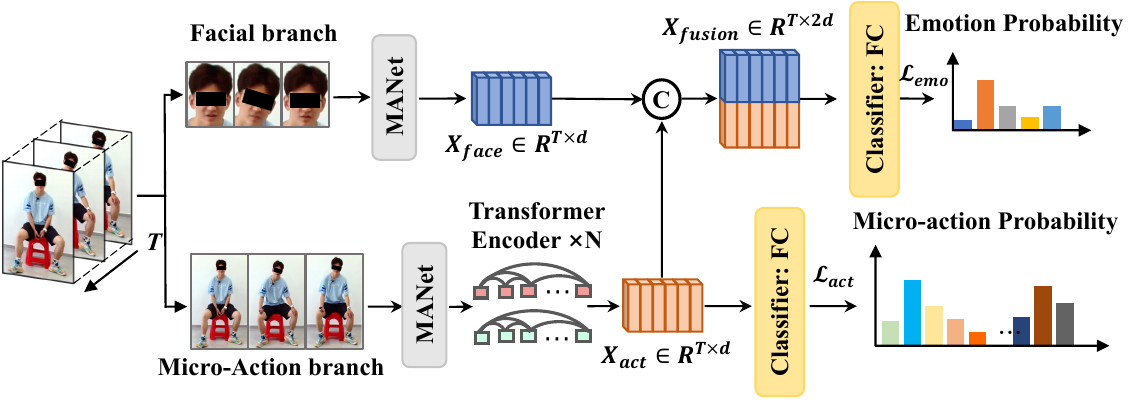}
\caption{MANet-based network architecture for emotion recognition. Both the facial and micro-action branches consistently adopt the MANet as the basic network unit. 
For multi-task optimization, we consider two losses $\mathcal{L}_{emo}$ and $\mathcal{L}_{act}$ to simultaneously optimize emotion recognition and micro-action occurrence.}
\vspace{-1.2em}
\label{fig:method_ear} 
\end{figure}

\textbf{Data Preparation.} 
We collect video instances that capture participants' emotion states to construct the MA-52-Pro dataset. 
The dataset consists of 7,818 instances containing 29,503 micro-action labels, spanning a total duration of 3,876 minutes. The emotion annotations are based on holistic participant data, including actions, expressions, speech, and tone. Each video instance manifests between 1 and 15 micro-actions with durations ranging from 5s to beyond 100s. Compared to existing bodily emotion datasets \cite{liu2021imigue,chen2023smg}, MA-52-Pro offers advantages such as high resolution data, diverse subjects, spontaneous emotional responses, and comprehensive annotations as well as MA-52. In contrast to iMiGUE~\cite{liu2021imigue} and SMG~\cite{chen2023smg}, which categorize emotions dichotomously (\ie, positive/negative and relaxed/stressed), the present dataset includes five emotional categories (\ie, joy, sadness, surprise, fear, and anger). 

\textbf{Method Pipeline.} 
As depicted in Figure~\ref{fig:method_ear}, we use the developed MANet as a basic module and introduce a unified MANet-based framework to simultaneously manage facial and micro-action branches for emotion recognition. In the facial branch, a pre-trained face detector~\cite{king2009dlib} is used to detect and crop facial regions and subsequently obtain facial features. After encoding by MANet, we obtain the final facial feature $\mathbf{X}_{face} \in \mathbb{R}^{T\times d}$. 
For the micro-action branch, the video input is processed by MANet. Since the extended MA-52-Pro dataset considers the simultaneous occurrence of multiple micro-actions, a two-layer transformer encoder~\cite{vaswani2017attention} is incorporated to detect long-term temporal associations among micro-actions. Ultimately, we obtain the micro-action features $\mathbf{X}_{act} \in \mathbb{R}^{T\times d}$. $\mathbf{X}_{act}$ is averaged along the timeline and then mapped to the FC-based micro-action classifier for multi-label micro-action prediction. 
The combination of $[\mathbf{X}_{face}, \mathbf{X}_{act}]$ is fed into the FC-based emotion classifier. The two classifiers are formulated as follows:
\begin{eqnarray}
\begin{aligned}
\left\{\begin{array}{l}
{\mathbf{A}} = FC(AvgPool(\mathbf{X}_{act}))\in \mathbb{R}^{1\times N^F_A}, \\
{\mathbf{E}} = FC(AvgPool([\mathbf{X}_{face};\mathbf{X}_{act}]))\in \mathbb{R}^{1\times N^E_C},
\end{array}\right.
\end{aligned}
\label{eq:emo}
\end{eqnarray}
where $N^F_A$ and $N^E_C$ denote the number of micro-action categories and emotion categories, respectively.

\textbf{Method Optimization.}
We leverage both micro-action and emotion indicators to construct two distinct optimization objectives, \ie, emotion optimization and multilabel micro-action optimization. The emotional loss $\mathcal{L}_{emo}$ denotes a normal cross-entropy loss designed for emotion prediction, and the actionness loss $\mathcal{L}_{act}$ represents a binary cross-entropy loss designed for multilabel micro-action prediction. The optimization is formulated as 
$\mathcal{L} = {L}_{act}+\mathcal \beta \mathcal{L}_{emo}$, 
where $\beta$ indicates a hyperparameter to balance the two losses. $\beta$ is empirically set to 0.03.

\begin{table}[t!]
\tiny
\renewcommand\arraystretch{1.3}
\centering
\caption{Performance comparison for emotion recognition on the MA-52-Pro. The best results are highlighted with \textbf{bold}.}
\resizebox{1.0\linewidth}{!}{
\begin{tabular}{l|cccc|c}
\hline
Method & Acc & F1$_\emph{{weight}}$ & UF1 & UAR &mAP \\ \hline
VAANet \cite{zhao2020end} &54.63 &48.43 &32.27 &32.06 &- \\
Emotion-FAN \cite{meng2019frame} &56.81 &53.89 & 37.42 &36.71 &-\\ \hline
Face Branch & 60.01 & 55.23 & 38.38 & 38.18 & 5.51 \\
Micro-Action Branch & 58.86 & 55.23 & 40.30 & 41.38 & 15.35 \\ 
\textbf{Full model} & \textbf{62.38} & \textbf{58.49} & \textbf{41.75} &\textbf{42.28} &\textbf{15.40} \\ \hline
\end{tabular}
}
\label{tab:emo-cmp}
\vspace{-1.2em}
\end{table}

\begin{figure*}[!t]
\centering
\includegraphics[width=1.0\linewidth]{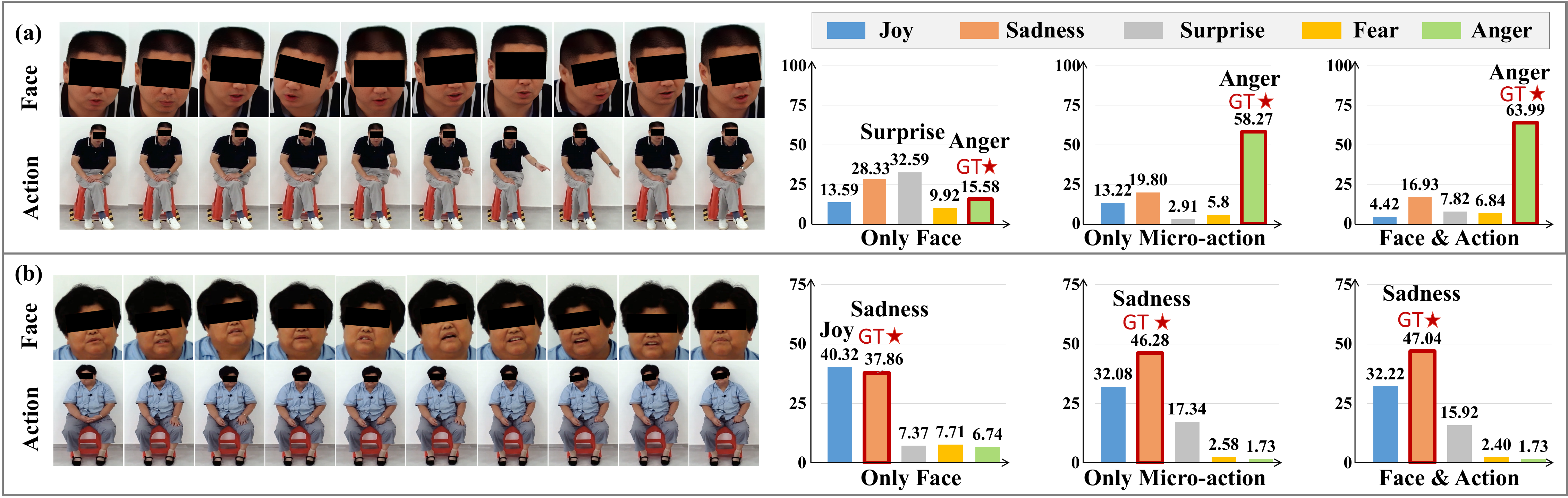}
\vspace{-2.0em}
\caption{Prediction results of the {only Face} branch, the {only Micro-action} branch, and the full model from the MA-52-Pro dataset. Histograms depict the probability distribution of the emotions. Herein, we present two cases where using only facial expressions is insufficient to distinguish ``anger'' from ``surprise'' and ``joy'' from ``sadness''. By leveraging micro-action cues, MAENet successfully predicts the correct category.}
\vspace{-1.2em}
\label{fig:emo-vis1} 
\end{figure*}

\begin{figure*}[t!]
\centering
\includegraphics[width=1.0\linewidth]{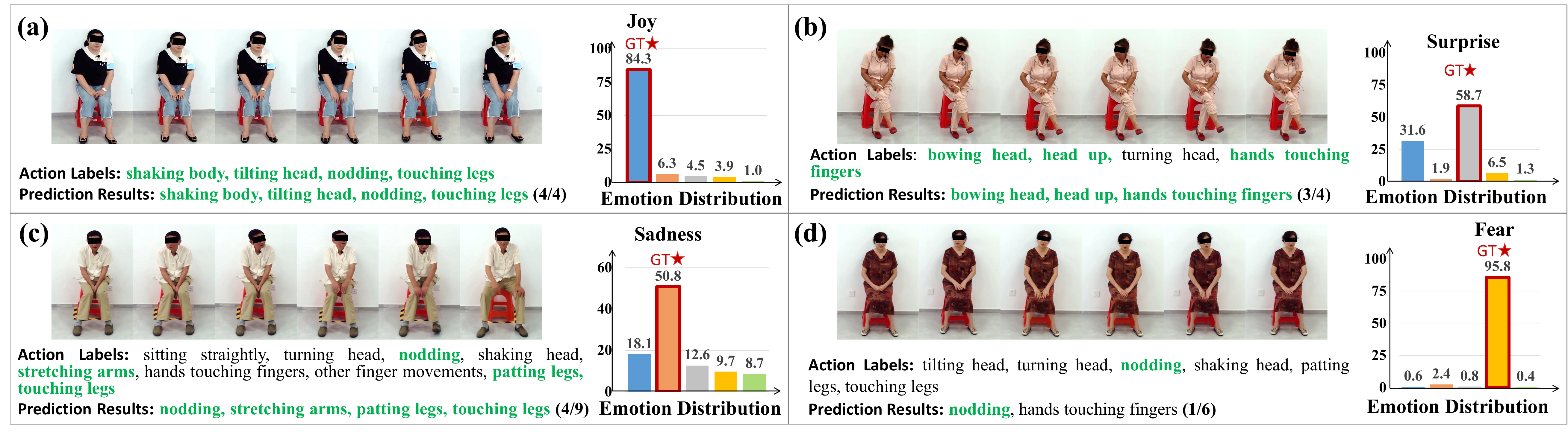}
\vspace{-2.0em}
\caption{Micro-action and emotion prediction results of four samples from the MA-52-Pro dataset. We list the predicted micro-action results and the ground-truth labels. All the emotion labels are annotated based on the participant's interview content and SCL-90 test result. The histogram plots the probability distribution of the emotions. The proposed method performs well on emotion recognition. 
Detecting and distinguishing massive simultaneous micro-actions in long videos remains a challenge. The proposed method identifies key micro-actions and successfully predicts the correct emotion category.}
\vspace{-1.2em}
\label{fig:emo-vis2} 
\end{figure*}

\begin{figure}[t!]
\centering
\includegraphics[width=1.0\linewidth]{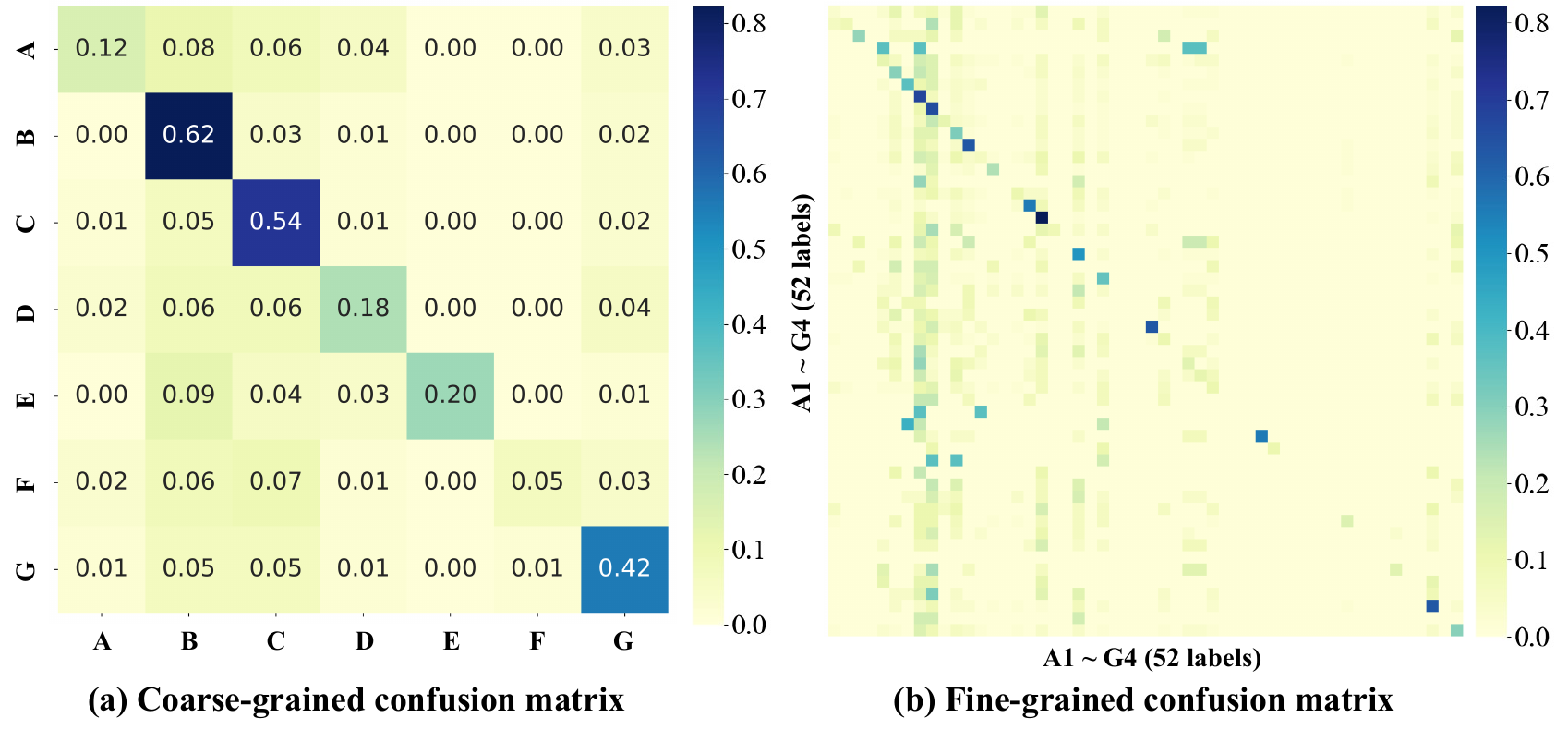}
\vspace{-1.0em}
\caption{Predicted confusion matrices of fine-grained and coarse-grained categories of micro-action categories on the MA-52-Pro test set. 
MA-52-Pro contains multiple simultaneous micro-actions. 
Although this method can easily learn the micro-action co-occurrences from the head, and upper limb (\ie, labels B and C in Fig.~\ref{fig:matrix2} a), whereas it is still challenging to detect multi-action co-occurrences from the body parts of body, lower limb, and head-hand (\ie, labels A, D, and F in Fig.~\ref{fig:matrix2} a).} 
\vspace{-1.2em}
\label{fig:matrix2} 
\end{figure}

\subsection{Experiment Analysis of Application}
Comparative analyses are conducted between our approach and emotion recognition methods \cite{zhao2020end,meng2019frame}. Additionally, we perform ablation studies that investigate both facial and micro-action branches. The experimental results are detailed in Table \ref{tab:emo-cmp}. Metrics such as Acc, F1$_{weight}$ \cite{shen2021dialogxl}, UF1 \cite{WANG202219,li2022deep} and UAR~\cite{chen2022block} are widely used for emotion evaluation. Mean Average Precision (mAP) denotes the average multi-label action classification accuracy for each video instance. For the implementation details, we empirically sample $T$=16 frames from each video instance, set the loss hyperparameter to $\beta$=0.03, and set the batch size to 3. The additional parameters are set the same as the ones in MANet.

\textbf{Quantitative Analysis.} 
{1) Emotion recognition:} As listed in Table \ref{tab:emo-cmp}, the proposed method achieves the best performances on Acc, F1$_{weight}$, UF1, and UAR metrics, reaching 62.38\%, 58.49\%, 41.75\%, and 42.88\%, respectively, which demonstrates the superiority of the proposed method for emotion analysis. 
{2) Multi-label micro-action recognition:} Due to the instances in MA-52-Pro containing multiple micro-action occurrences, the model is inevitably insufficient to predict all the action categories, especially in the case of massive simultaneous micro-actions occurring in long videos. It achieves 15.40\% on the micro-action recognition mAP, which is an acceptable result. 
{3) Facial and micro-action branches:} ``Only Micro-action" performs slightly worse than ``Only Face". Thus, the study of micro-action based emotion recognition is still challenging and deserves further attention and research. 
{4) Confusion matrices of the micro-action prediction results}: The MA-52-Pro dataset contains several simultaneous micro-actions. Ideally, the confusion matrices in Figure~\ref{fig:matrix2} would prominently highlight the co-occurrence phenomena of multiple actions. The proposed method successfully learns to identify such co-occurrences from the head, and upper limb (\ie, labels B and C in Figure~\ref{fig:matrix2} (a)). However, it still encounters challenges in detecting co-occurrences involving full-body, lower limbs, and head-hand actions (\ie, coarse-grained labels A, D and F in Figure~\ref{fig:matrix2} (a)).

\begin{table*}[t!]
\centering
\begin{minipage}{1.0\linewidth}
\centering
\caption{
The statistics of action label numbers of the MA-52-Pro dataset and the ablation studies of lower limb micro-actions for emotion recognition on the dataset.}
\resizebox{1.0\linewidth}{!}{
\begin{tabular}{c|ccccccc|c}
\Xhline{1pt}
Label & Body (A) & Head (B) & Upper limb (C) & \cellcolor{gray!25}Lower limb (D) & Body-hand (E) & Head-hand (F) & \cellcolor{gray!25}Leg-hand (G) & All \\ \hline
Number &   1,492   &  1,4064 & 6,988 & \cellcolor{gray!25}3,290  & 422  & 1,158 & \cellcolor{gray!25}2,089 & 29,503   \\ \hline
\end{tabular}}
\label{tab:stat_ma52pro}
\end{minipage}
\vspace{0.2cm}
\begin{minipage}{1.0\linewidth}
\centering
\resizebox{1.0\linewidth}{!}{
\begin{tabular}{c|c|ccccc|ccc|c}
\hline
Metric & Acc\_Avg &Acc\_Joy &Acc\_Sadness &Acc\_Surprise &Acc\_Fear &Acc\_Anger& F1$_{weight}$ & UF1 & UAR & mAP \\ \hline
w/o. lower limb labels D\&G &\underline{60.65} &  \underline{88.35}   &  \underline{53.32} &  \textbf{47.08}   &   \textbf{9.09}  & \textbf{7.40}& \underline{56.90} & \textbf{41.75}&\underline{41.05} &\textbf{16.20} \\ 
w. full labels  &  \textbf{62.38}  &\textbf{90.55} & \textbf{57.30} & \underline{45.62}& \textbf{9.09} &\underline{6.24} &   \textbf{58.49}  &  \textbf{41.75}   &   \textbf{42.28}  & \underline{15.40} \\ \Xhline{1pt}
\end{tabular}}
\label{tab:Ablation_ma52pro}
\end{minipage}
\end{table*}

\textbf{Qualitative Visualization.} Figure~\ref{fig:emo-vis1} provides visual examples from the Face-only branch, the Micro-action-only branch, and the full model. Two illustrative cases show the limitations of relying solely on facial expressions to distinguish between ``anger'' and ``surprise'' or ``joy'' and ``sadness.'' 
By utilizing micro-action cues, MAENet accurately predicts the correct emotion categories, underlining the robustness of the model.
Figure~\ref{fig:emo-vis2} showcases four examples of both micro-action and emotion recognition. Our method performs commendably well in the area of emotion recognition. In terms of multi-label micro-action recognition, it particularly excels in handling video instances characterized by a limited number of micro-actions, as displayed in Figures~\ref{fig:emo-vis2} (a) and (b). 
However, predicting multiple simultaneous micro-actions in extended videos or large sets of actions remains a challenging task, as shown in Figures~\ref{fig:emo-vis2} (c) and (d). Nevertheless, our method can identify key micro-actions and successfully predict the correct emotion category. 

\textbf{Necessity of Lower Limb Labels for Emotion Recognition.}
We further conduct the experiment on the MA-52-Pro dataset to verify the effectiveness of lower limb micro-actions for emotion recognition. 
As shown in Table~\ref{tab:stat_ma52pro}, we report the statistics of the actions of each part in the MA-52-Pro dataset. There are 29,503 action labels under 7 coarse-grained labels including Body (A), Head (B), Upper limb (C), Lower limb (D), Body-hand (E), Head-hand (F), and Leg-hand (G).  
In fact, the lower limb part contains both ``Lower limb (D)'' and ``Leg-hand (G)'' micro-actions in our MA-52-Pro dataset, reaching 5,379 labels and 18.23\% of the total. 
Figure~\ref{fig:emo-act_label} (a) shows that lower limb micro-actions account for a non-negligible proportion of micro-actions in the five categories of emotions, reaching 20\% for Joy, 16\% for Sadness, 18\% for Surprise, 16\% for Fear, and 13\% for Anger. 
We then carry out an ablation study of the model that does not use the lower limb part of the micro-actions for emotion recognition. 
From the experimental results in Table~\ref{tab:stat_ma52pro}, 
the model without the lower limb part shows performance drops in Acc\_Avg (1.73\%$\downarrow$), F1$_{weight}$ (1.59\%$\downarrow$), and UAR (1.23\%$\downarrow$). In addition, we find that the mAP index shows an increase after removing lower limb micro-actions, indicating that the recognition of lower limb micro-actions is still a challenge.

\begin{figure}[!t]
\centering
\includegraphics[width=1\linewidth]{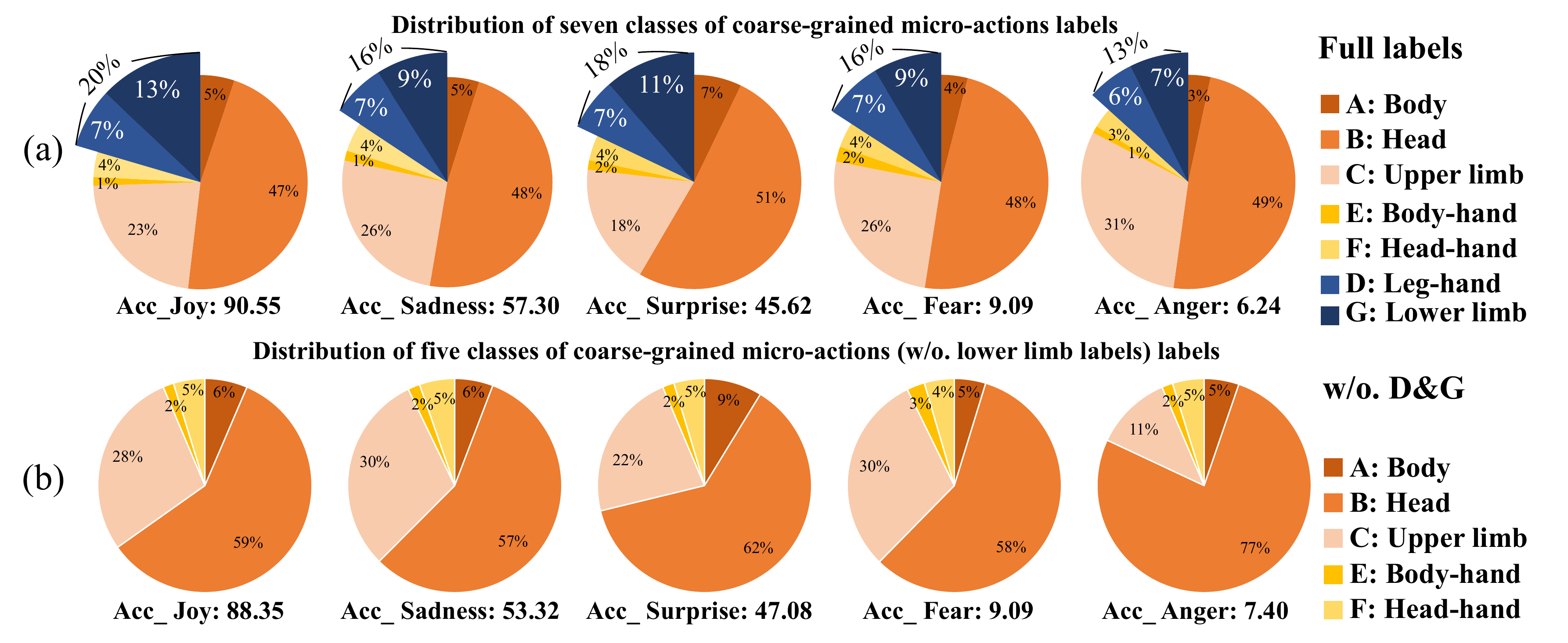}
\caption{Percentage of micro-action labels in different emotion categories.
(a) w. full micro-action labels; (b) w/o. lower limb micro-action labels D\&G. The lower limb micro-actions account for a non-negligible proportion of micro-actions for emotion recognition.}
\label{fig:emo-act_label} 
\end{figure}

\section{Further Research Direction}
\label{sec:future}
MAR is emerging as a research area of great interest today, which has shown great potential in a variety of application scenarios. 
Here, we provide an in-depth discussion on the further research directions of micro-action recognition: 
\textbf{1) Composite MAR:} In the real world, people's behaviors and actions often involve multiple co-occurring actions, not just isolated micro-action. 
Thus, composite micro-actions offer more true human behavior patterns in complex real-world scenarios. Future research in MAR should consider composite micro-action.  
\textbf{2) Multimodal MAR:}
Multimodal micro-action is a research direction that integrates information from multiple perceptual modalities for a more comprehensive and in-depth understanding and recognition of human micro-actions, such as the modalities of audio and depth data. 
\textbf{3) Micro-expression and Micro-action (ME\&MA) Incorporation:}
Both micro-expressions and micro-actions reflect a person's true feelings. 
Future research should jointly exploit micro-expressions and micro-actions to thoroughly improve the understanding of linkage.
It is expected to bring innovations in psychology, emotional intelligence technology, and human-computer interaction. 
\textbf{4) Micro-action Generation (MAG):}
AI-generated content (AIGC) can significantly increase the variety and volume of micro-action data, overcoming the limitations of expensive collection and annotation associated with traditional data collection. 
With further investigation, the generated micro-actions can be utilized in the development of real-world human-centric applications. 
\textbf{5) More Challenging Realistic Situations:}
Micro-action recognition still faces several challenges in realistic situations.
Capturing the micro-actions of people in the wild cannot avoid noise, including background clutter, occlusion, lighting variations, and camera shake. These noise factors can significantly affect micro-motion detection and recognition accuracy. Future work should pay more attention to exploring the generalization and robustness of the method in realistic situations.

\section{Conclusion}
\label{sec:conclusion}
To make progress in this particular field of research, we explore video-based human micro-action recognition and compile an expansive, class-rich, and whole-body Micro-Action-52 (MA-52) dataset. This study 
scrutinizes prevalent generic action recognition methods and 
introduces a CNN network architecture named MANet. The method integrates SE with the TSM into the established ResNet network to capture nuanced spatio-temporal variations within videos. Experimental results demonstrate both the reliability of the assembled dataset and the effectiveness of the proposed method. This study employs the proposed micro-action recognition method for emotion recognition task, performing multi-task learning for multi-label micro-action recognition and emotion recognition tasks, which verifies the proficiency of MANet in emotion analysis applications. Micro-action recognition serves as a highly practical technology that provides numerous opportunities in various real-world applications, thus opening up various avenues for future academic exploration.

\section*{Acknowledgments}
We would like to thank the ethics committee of the Institute of Artificial Intelligence, Hefei Comprehensive National Science Center, for supervising the collection and usage of the dataset. As well, we would like to thank all the participants interviewed for the dataset. Prior to data collection, we informed each participant of the requirement for data collection and obtained their signed consents for academic research purposes. 

\ifCLASSOPTIONcaptionsoff
\newpage
\fi

{\small
\bibliographystyle{IEEEtran}

\bibliography{IEEE}
}

\end{document}